\documentclass[10pt,twocolumn,letterpaper]{article}

\usepackage{cvpr}              % To produce the CAMERA-READY version
\usepackage{graphicx}
\usepackage{amsmath}
\usepackage{amssymb}
\usepackage{booktabs}

\usepackage{xcolor,colortbl}
\definecolor{lgreen}{RGB}{236, 255, 201}
\definecolor{nvgreen}{RGB}{118, 185, 0}

\usepackage[pagebackref=false,breaklinks=true,letterpaper=true,bookmarks=false,colorlinks,citecolor=nvgreen,linkcolor=nvgreen]{hyperref}

\usepackage{times}
\usepackage{epsfig}
\usepackage{graphicx}
\usepackage{amsmath}
\usepackage{amssymb}
\usepackage{multirow}
\usepackage{microtype}
\usepackage[normalem]{ulem}
\useunder{\uline}{\ul}{}
\usepackage{color}
\usepackage{nicefrac}
\usepackage{comment}
\usepackage{enumitem}
\usepackage{adjustbox}
\usepackage{mathabx}
\usepackage{tabu}
\usepackage{booktabs}
\usepackage{diagbox}
\usepackage{cuted}
\usepackage{capt-of} 

\usepackage{amssymb}

\usepackage{pifont}
\usepackage{algorithm}
\usepackage{algpseudocode} 
\usepackage[capitalize]{cleveref}
\crefname{section}{Sec.}{Secs.}
\Crefname{section}{Section}{Sections}
\Crefname{table}{Table}{Tables}
\crefname{table}{Tab.}{Tabs.}

\def\cvprPaperID{1476} % *** Enter the CVPR Paper ID here
\def\confName{CVPR}
\def\confYear{2022}

\newcommand{\xmark}{\ding{55}}%
\newcommand{\cmark}{\text{\ding{51}}}

\usepackage{overpic}
\usepackage{enumitem} %< control spacing in itemize/enumerate/...
\usepackage{overpic} %< add raw math symbols to figures
\usepackage{color}
\definecolor{turquoise}{cmyk}{0.65,0,0.1,0.3}
\definecolor{purple}{rgb}{0.65,0,0.65}
\definecolor{dark_green}{rgb}{0, 0.5, 0}
\definecolor{orange}{rgb}{0.8, 0.6, 0.2}
\definecolor{red}{rgb}{0.8, 0.2, 0.2}
\definecolor{darkred}{rgb}{0.6, 0.1, 0.05}
\definecolor{blueish}{rgb}{0.0, 0.3, .6}
\definecolor{light_gray}{rgb}{0.7, 0.7, .7}
\definecolor{pink}{rgb}{1, 0, 1}
\definecolor{greyblue}{rgb}{0.25, 0.25, 1}

\usepackage{blindtext}

\renewcommand{\paragraph}[1]{\vspace{1em}\noindent\textbf{#1}.}
\begin{document}
\title{A-ViT: Adaptive Tokens for Efficient Vision Transformer}
% \title{A-ViT: Adaptive Tokens for Efficient Vision Transformer}

\author{Hongxu Yin \hspace{0.3cm} 
Arash Vahdat \hspace{0.3cm}
Jose M. Alvarez \hspace{0.3cm} 
Arun Mallya \hspace{0.3cm} 
Jan Kautz \hspace{0.3cm} 
Pavlo Molchanov\\
NVIDIA\\
{\tt\small \{dannyy, avahdat, josea, amallya, jkautz, pmolchanov\}@nvidia.com}
% For a paper whose authors are all at the same institution,
% omit the following lines up until the closing ``}''.
% Additional authors and addresses can be added with ``\and'',
% just like the second author.
% To save space, use either the email address or home page, not both
% \and
% Second Author\\
% Institution2\\
% First line of institution2 address\\
% {\tt\small secondauthor@i2.org}
}

% \title{[CVPR2022] Official LaTeX Template}

% \author{Andrea Tagliasacchi\\
% Google Research \& University of Toronto\\
% {\tt\small taglia@google.com}
% % For a paper whose authors are all at the same institution,
% % omit the following lines up until the closing ``}''.
% % Additional authors and addresses can be added with ``\and'',
% % just like the second author.
% % To save space, use either the email address or home page, not both
% % \and
% % Second Author\\
% % Institution2\\
% % First line of institution2 address\\
% % {\tt\small secondauthor@i2.org}
% }

\maketitle

\begin{strip}\centering
\vspace{-15mm}
\includegraphics[width=\textwidth,clip]{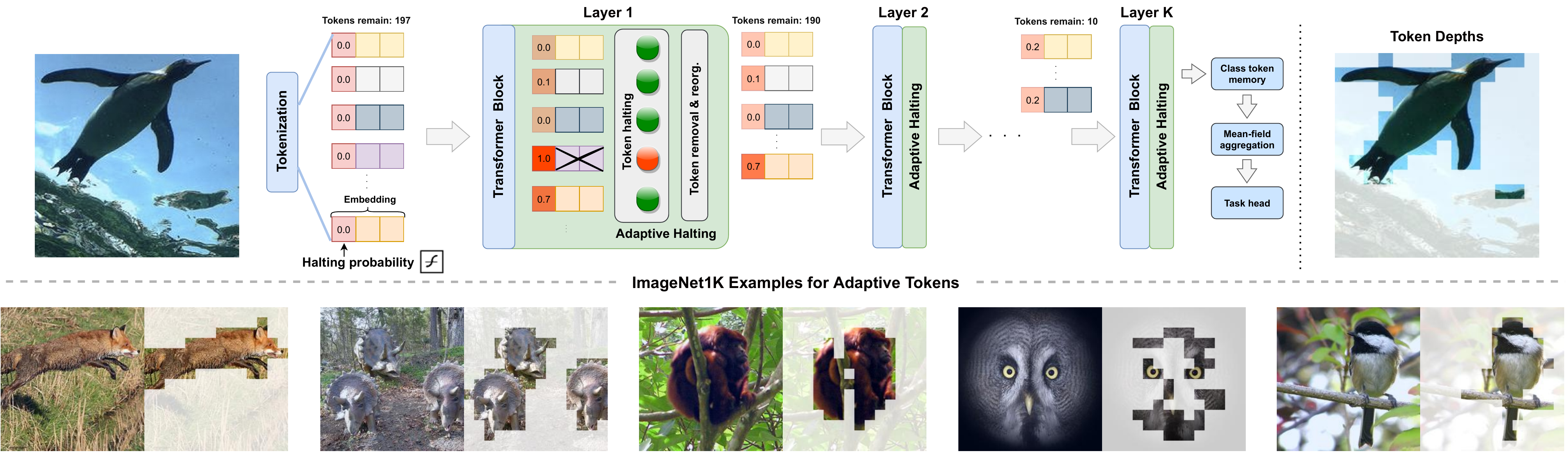}
\captionof{figure}{
We introduce A-ViT, a method to enable \textit{\textbf{adaptive token}} computation for vision transformers. We augment the vision transformer block with adaptive halting module that computes a halting probability per token. The module reuses the parameters of existing blocks and it borrows a single neuron from the last dense layer in each block to compute the halting probability, imposing no extra parameters or computations. A token is discarded once reaching the halting condition. Via adaptively halting tokens, we perform dense compute only on the active tokens deemed informative for the task. As a result, successive blocks in vision transformers gradually receive less tokens, leading to faster inference. Learnt token halting vary across images, yet align \textbf{\textit{surprisingly well}} with image semantics (see examples above and more in Fig.~\ref{fig:all_images}). This results in immediate, out-of-the-box inference speedup on off-the-shelf computational platform. 
%\AV{the rest is too much:} Token halting inspired by adaptive computation time (ACT)~\cite{graves2016adaptive}, adapted to mean-field formulation of class token and halting of patch tokens to save computation during inference.
% \JM{Last sentence seems unfinished} 
% Please see token Link to the source in the fig/teaser.tex
%https://drive.google.com/file/d/1xmj4PGDJqxvFumyTIki4PPtCjWK1_jgM/view?usp=sharinga
}
\label{fig:teaser}
\end{strip}
\begin{abstract}
\vspace{-2mm}
We introduce A-ViT, a method that adaptively adjusts the inference cost of vision transformer (ViT) for images of different complexity. A-ViT achieves this by automatically reducing the number of tokens in vision transformers that are processed in the network as inference proceeds. We reformulate Adaptive Computation Time (ACT \cite{graves2016adaptive}) for this task, extending halting to discard redundant spatial tokens. The appealing architectural properties of vision transformers enables our adaptive token reduction mechanism to speed up inference without modifying the network architecture or inference hardware. We demonstrate that A-ViT requires no extra parameters or sub-network for halting, as we base the learning of adaptive halting on the original network parameters. We further introduce distributional prior regularization that stabilizes training compared to prior ACT approaches. On the image classification task (ImageNet1K), we show that our proposed A-ViT yields high efficacy in filtering informative spatial features and cutting down on the overall compute. The proposed method improves the throughput of DeiT-Tiny by $62\%$ and DeiT-Small by $38\%$ with only $0.3\%$ accuracy drop, outperforming prior art by a large margin\let\thefootnote\relax\footnotetext{Project page at \href{https://a-vit.github.io/}{https://a-vit.github.io/}.}.%
\vspace{-4mm}
%and allocating more compute wisely only towards such regions, 
\end{abstract}
\section{Introduction}
\label{sec:intro}

%DL and transformers
% old version
% Deep learning (DL) models learn a non linear function to transform input to the required output. They are the core of recent breakthrough in computer vision and significantly advanced image understanding (classification, detection, segmentation). Practical applications such as smart driver assistance and self-driving is now possible. Architecture of the DL models improves with advances in model design, theoretical and empirical studies. If few years ago convolutional neural networks where dominating the field, then now visual transformers are popular among researchers. Transformers split the image into a series of ordered patches and perform inter and intra calculations to solve the task. 

% new version
Transformers have emerged as a popular class of neural network architecture that computes network outputs using highly expressive attention mechanisms. Originated from the natural language processing (NLP) community, they have been shown effective in solving a wide range of problems in NLP, such as machine translation, representation learning, and question answering~\cite{vaswani2017attention,devlin2018bert,radford2018improving,jiao2019tinybert,brown2020language}. Recently, vision transformers have gained an increasing popularity in the vision community and they have been successfully applied to a broad range of vision applications, such as image classification~\cite{dosovitskiy2020image,touvron2021training,yuan2021tokens,graham2021levit,liu2021swin,wang2021pyramid}, object detection~\cite{carion2020end,dai2021up,sun2021rethinking}, image generation~\cite{jiang2021transgan,hudson2021generative}, and semantic segmentation~\cite{xie2021segformer,li2021panoptic}. The most popular paradigm remains when vision transformers form tokens via splitting an image into a series of ordered patches and perform inter-/intra-calculations between tokens to solve the underlying task. Processing an image with vision transformers remains computationally expensive, primarily due to the quadratic number of interactions between tokens~\cite{yang2021nvit,tambe2020edgebert,rao2021dynamicvit}. Therefore, deploying vision transformers on data processing clusters or edge devices is challenging amid significant computational and memory resources.

%why we need to save compute?
% old version
% DL models perform a forward pass of the data through the model to emit the output. This requires computing multiple linear and nonlinear operations with different complexity. Applying DL models to solve tasks at hand requires to run inference for every received image. Computational clusters and edge devices spend significant computational and memory resources when use them in practise. Both types of hardware will benefit from reducing the complexity of computation and algorithmic speed-ups during the inference.

% new version
% Processing an image with vision transformers remains computationally expensive, primarily due to the quadratic number of interactions between tokens~\cite{yang2021nvit,tambe2020edgebert,rao2021dynamicvit}. Therefore, deploying vision transformers on either data processing clusters or edge devices is challenging as they require significant computational and memory resources. Both types of hardware can significantly benefit from reducing the complexity of computation and algorithmic speed-ups during inference.

The main focus of this paper is to study how to automatically adjust the compute in visions transformers as a function of the complexity of the input image. Almost all mainstream vision transformers have a fixed cost during inference that is independent from the input. However, the difficulty of a prediction task varies with the complexity of the input image. For example, classifying a car versus a human from a single image with a homogeneous background is relatively simple; while differentiating between different breeds of dogs on a complex background is more challenging. Even within a single image, the patches that contain detailed object features are far more informative compared to those from the background. Inspired by this, we develop a framework that adaptively adjusts the compute used in vision transformers based on the input.

The problem of input-dependent inference for neural networks has been studied in prior work. 
Graves~\cite{graves2016adaptive} proposed adaptive computation time (ACT) to represent the output of the neural module as a mean-field model defined by a halting distribution. Such formulation relaxes the discrete halting problem to a continuous optimization problem that minimizes an upper bound on the total compute. Recently, stochastic methods were also applied to solve this problem, leveraging geometric-modelling of exit distribution to enable early halting of network layers~\cite{banino2021pondernet}. Figurnov \etal~\cite{figurnov2017spatially} proposed a spatial extension of ACT that halts convolutional operations along the spatial cells rather than the residual layers. This approach does not lead to faster inference as high-performance hardware still relies on dense computations. However, we show that the vision transformer's uniform shape and tokenization enable an adaptive computation method to yield a direct speedup on off-the-shelf hardware, surpassing prior work in efficiency-accuracy tradeoff.

In this paper, we propose an input-dependent adaptive inference mechanism for vision transformers. A naive approach is to follow ACT, where the computation is halted for all tokens in a residual layer simultaneously. We observe that this approach reduces the compute by a small margin with an undesirable accuracy loss. To resolve this, we propose A-ViT, a spatially adaptive inference mechanism that halts the compute of different tokens at different depths, reserving compute for only discriminative tokens in a dynamic manner. Unlike point-wise ACT within convolutional feature maps~\cite{figurnov2017spatially}, our spatial halting is directly supported by high-performance hardware since the halted tokens can be efficiently removed from the underlying computation. Moreover, entire halting mechanism can be learnt using existing parameters within the model, without introducing any extra parameters. We also propose a novel approach to target different computational budgets by enforcing a distributional prior on the halting probability. We empirically observe that the depth of the compute is highly correlated with the object semantics, indicating that our model can ignore less relevant background information (see quick examples in Fig.~\ref{fig:teaser} and more examples in Fig.~\ref{fig:all_images}). Our proposed approach significantly cuts down the inference cost -- A-ViT improves the throughput of DEIT-Tiny by $62\%$ and DEIT-Small by $38\%$ with only $0.3\%$ accuracy drop on ImageNet1K. 

Our main contributions are as follows: 
% \PM{we can probably skip contributions}
\begin{itemize}[noitemsep,nosep]
    \item We introduce a method for input-dependent inference in vision transformers that allows us to halt the computation for different tokens at different depth. 
    \item We base learning of adaptive token halting on the existent embedding dimensions in the original architecture and do not require extra parameters or compute for halting.
    \item We introduce distributional prior regularization to guide halting towards a specific distribution and average token depth that stabelizes ACT training. 
    \item We analyze the depth of varying tokens across different images and provide insights into the attention mechanism of vision transformer. 
    \item We empirically show that the proposed method improves throughput by up to $62\%$ on hardware with minor drop in accuracy. 
\end{itemize}

\section{Related Work}
\label{sec:related}

There are a number of ways to improve the efficiency of transformers including weight sharing across transformer blocks~\cite{lan2019albert}, dynamically controlling the attention span of each token~\cite{chen2019adaptive,tambe2020edgebert}, allowing the model to output the result in an earlier transformer block~\cite{zhou2020bert,schwartz2020right}, and applying pruning~\cite{yang2021nvit}. A number of methods have aimed at reducing the computationally complexity of transformers by reducing the quadratic interactions between tokens~\cite{tay2020sparse, katharopoulos2020transformers, choromanski2021rethinking, Kitaev2020Reformer, wang2020linformer}. We focus on approaches related to adaptive inference that depends on the input image complexity. A more detailed analysis of the literature is present in~\cite{han2021dynamic}. 

\noindent 
\textbf{Special architectures.} One way is to change the architecture of the model to support adaptive computations~\cite{fung2021jfb,yang2020resolution, leroux2018iamnn,chen2019neural,xia2021fully, guo2019dynamic, teerapittayanon2016branchynet, liu2018dynamic, rota2014neural, kontschieder2015deep, frosst2017distilling}. For example, models that represent a neural network as a fixed-point function can have the property of adaptive computation by default. Such models compute the difference to the internal state and, when applied over multiple iterations, converge towards the solution (desired output). For example, neural ordinary differential equations (ODEs) use a new architecture with repetitive computation to learn the dynamics of the process~\cite{dong2020towards}. Using ODEs requires a specific solver, is often slower than fix depth models and requires adding extra constraints on the model design. \cite{yang2020resolution} learns a set of classifiers with different resolutions executed in order; computation stops when confidence of the model is above the threshold. \cite{leroux2018iamnn} proposed a residual variant with shared weights and a halting mechanism.

\noindent
\textbf{Stochastic and reinforcement learning (RL) methods.} The depth of a residual neural network can be changed during inference by skipping a subset of residual layers. This is possible since residual networks have the same input and output feature dimensions and they are known to perform feature refinements iteratively. Individual extra models can be learned on the top of a backbone to change the computational graph. A number of approaches ~\cite{wang2018skipnet, wu2018blockdrop, lin2017runtime, odena2017changing} proposed to train a separate network via RL to decide when to halt. These approaches require training of a dedicated halting model and their training is challenging due to the high-variance training signal in RL. Conv-AIG~\cite{veit2018convolutional} learns conditional gating of residual blocks via Gumbel-softmax trick. \cite{verelst2020dynamic} extends the idea to spatial dimension (pixel level).   

\noindent 
\textbf{Adaptive inference in vision transformers.} With the increased popularity, researchers have very recently explored adaptive inference for vision transformers. DynamicViT~\cite{rao2021dynamicvit} uses extra control gates that are trained with the Gumbel-softmax trick to halt tokens and it resembles some similarities to Conv-AIG~\cite{veit2018convolutional} and \cite{verelst2020dynamic}. Gumbel-softmax-based relaxation solutions might be sub-optimal due to the difficulty of regularization, stochasticity of training, and early convergence of the stochastic loss, requiring multi-stage token sparsification as a heuristic guidance. In this work, we approach the problem from a rather different perspective, and we study how an ACT~\cite{graves2016adaptive}-like approach can be defined for spatially adaptive computation in vision transformers. We show complete viability to remove the need for the extra halting sub-networks, and we show that our models bring simultaneous efficiency, accuracy, and token-importance allocation improvements, as shown later.  

\section{A-ViT}

% We next explain A-ViT in details, where we formulate the problem, articulate token halting paradigm, design parameter-free halting networks, and impose distributional prior as training guidance. 

\begin{figure*}[t]
\centering
\resizebox{.95\linewidth}{!}{
\includegraphics[width=0.25\linewidth]{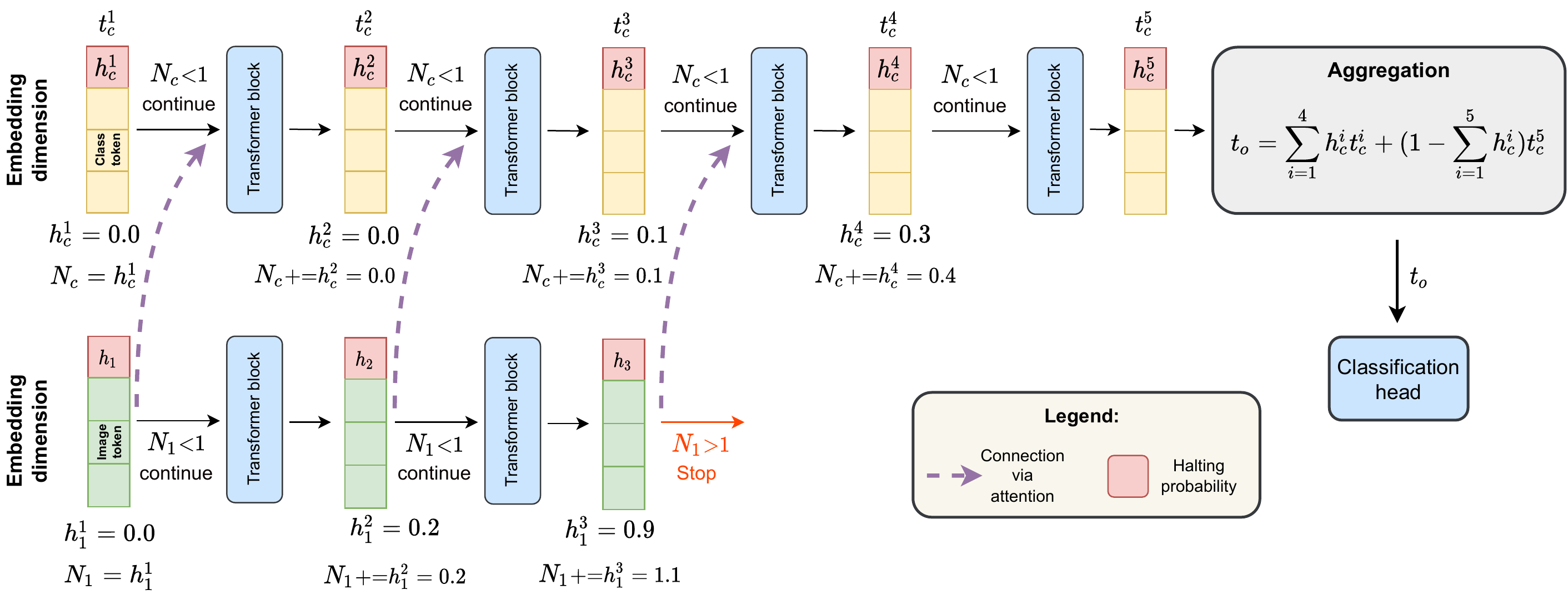}
}
\caption{An example of A-ViT: In the visualization, we omit (i) other patch tokens, (ii) the attention between the class and patch token and (iii) residual connections for simplicity. The first element of every token is reserved for halting score calculation, adding no computation overhead. We denote the class token with a subscript $c$ as it has a special treatment. Each token indexed by $k$ has a separate $N_k$ accumulator and stop at different depths. Unlike standard ACT, the mean-field formulation is applied only to the classification token, while other tokens contribute to the class token via attention. This allows adaptive token calculation without the aggregation of image/patch tokens. 
% \AV{Aggregation formula in right side does not match Eq.~\ref{eq:aggregation}: $p_c$ is used in the formula but $h_c$ is used in the figure.} \yin{seems we unroll $p_c$ here and explicitly write it out all in $h_c$ by Pavlo.}
}
\label{fig:main_algo}
\end{figure*}

%Consider a vision transformer network that makes a prediction on input image $x\in \mathcal{R}^{C\times H \times W}$ ($C$, $H$, and $W$ represent channel, height, and width respectively) through:
Consider a vision transformer network that takes an image $x\in \mathcal{R}^{C\times H \times W}$ ($C$, $H$, and $W$ represent channel, height, and width respectively) as input to make a prediction through:
\begin{equation}
    y = \mathcal{C} \circ \mathcal{F}^{L} \circ \mathcal{F}^{L-1} \circ ... \circ \mathcal{F}^1 \circ \mathcal{E}(x),
\end{equation}
where the encoding network $\mathcal{E}(\cdot)$ tokenizes the image patches from $x$ into the positioned tokens $t\in \mathcal{R}^{K \times E}$, $K$ being the total number of tokens and $E$ the embedding dimension of each token. 
% \AV{I have a hard time understanding what the last sentence means.}\JK{same. Also use: $ \#\text{token}$ instead of the other way around}. 
$\mathcal{C}(\cdot)$ post-processes the transformed class token after the entire stack, while the $L$ intermediate transformer blocks $\mathcal{F}(\cdot)$ transform the input via self-attention. 
% We aim to simplify the necessary compute before feeding tokens into $\mathcal{F}(\cdot)$ by adaptively adjusting the incoming tokens per target layer block. For simplicity, we index each token within the $K$ original token set as $k$. 
% We leave $\mathcal{F}(\cdot)$ intact so that no modification is needed to alter the transformer blocks.
Consider the transformer block at layer $l$ that transforms all tokens from layer $l-1$ via: 
% \AV{transformers are set encoders, they don't operate on individual tokens. Each block transforms all tokens together. That's why I changed $t_{k}^{l}$ to $t_{1:K}^{l}$ below.}\yin{agree}
\begin{equation}
    t_{1:K}^{l} = \mathcal{F}^{l} (t_{1:K}^{l-1}),
\label{eqn:main}
\end{equation}
where $t_{1:K}^{l}$ denotes all the $K$ updated token, with $t_{1:K}^{0}=\mathcal{E}(x)$. Note that the internal computation flow of transformer blocks $\mathcal{F}(\cdot)$ is such that the number of tokens $K$ can be changed from a layer to another. This offers out-of-the-box computational gains when tokens are dropped due to the halting mechanism. Vision transformer~\cite{dosovitskiy2020image, touvron2021training} utilizes a consistent feature dimension $E$ for all tokens throughout layers. This makes it easy to learn and capture a \textit{global} halting mechanism that monitors all layers in a joint manner. This also makes halting design easier for transformers compared to CNNs that require explicit handling of varying architectural dimensions, \eg, number of channel, at different depths. 
% Yet such dropping has to be done smartly, and adaptively by intuition -- given a token's one-to-one bond to each spatial location, it can represent a critical feature for one image, or merely a single-colored background for another image. Another important aspect of tokens is that in the most widely adopted regime, vision transformer utilizes a consistent dimension for $t^{l}, \forall l = 1, 2, ..., L$ throughout layers. This offers ease to learn and capture a \textit{global} halting mechanism that oversees all layers in a joint manner, imposing less design efforts as to halting CNNs that require explicit handling for varying architectural dimensions, \eg, number of channels, at different depths. 

% with the dimension of token number\JK{sounds odd} intact during its internal inference via reshaping. 
% This allows for an explicit efficiency boost if we can dynamically adjust the token numbers on-the-fly 
% \JK{this seems counter-intuitive: we highlight that it's great that there is a consistent number of tokens, but we will dynamically adjust it! Need to phrase this differently. I think this is where you want to describe how you can drop tokens and achieve a speed-up.}

To halt tokens adaptively, we introduce an input-dependent halting score for each token as
% through forming the halting score for each token
% \JM{Do not understand}\JK{it seems like a circular sentence...}, 
a halting probability $h_k^l$ for a token $k$ at layer $l$:
\begin{equation}
    h_k^l = H(t_k^l),
\end{equation}
where $H(\cdot)$ is a halting module. Akin to ACT~\cite{graves2016adaptive}, we enforce the halting score of each token $h^l_k$ to be in the range $0\leq h^l_k \leq 1$, and use accumulative importance to halt tokens as inference progresses into deeper layers. To this end, we conduct the token stopping when the cumulative halting score exceeds $1-\epsilon$: 
% \JK{I think the equation is wrong. The sum should be $\sum_{l=1}^{n}$, right?}
\begin{equation}
    N_{k} =\underset{ n \leq L}{\text{argmin }} \sum_{l=1}^{n} h^l_k \geq 1 - \epsilon,
\end{equation}
where $\epsilon$ is a small positive constant that allows halting after one layer. To further alleviate any dependency on dynamically halted tokens between adjacent layers, we mask out a token $t_k$ for all remaining depth $l>N_k$ once it is halted by (i) zeroing out the token value, and (ii) blocking its attention to other tokens, shielding its impact to $t^{l>N_k}$ in Eqn.~\ref{eqn:main}. We define $h_{1:K}^L=\mathbf{1}$ to enforce stopping at the final layer for all tokens. Our token masking keeps the computational cost of our training iterations similar to the original vision transformer's training cost. However, at the inference time, we simply remove the halted tokens from computation to measure the actual speedup gained by our halting mechanism. 
% \AV{A reviewer may ask why do you only zero out the masked tokens and why don't you remove those tokens from the computation during training similar to the test time. Can we say something about it?}

%It might seem as if extra learnable parameters are needed to construct $H(\cdot)$ as auxiliary networks, we observe that the existing parameters in vision transformers can already fulfill this task.
We incorporate $H(\cdot)$ into the existing vision transformer block by allocating a single neuron in the MLP layer to do the task. Therefore, we do not introduce any additional learnable parameters or compute for halting mechanism. More specifically, we observe that the embedding dimension $E$ of each token spares sufficient capacity to accommodate learning of adaptive halting, enabling halting score calculation as:
\begin{equation}
    H(t_k^l) = \sigma(\gamma \cdot t_{k, e}^l + \beta),
\end{equation}
% \begin{equation}
%      h_k^l = H(t_k^l) = \sigma(\gamma \cdot t_{k, e=0}^l + \beta),
% \end{equation}
where $t^l_{k, e}$ indicates the $e^{\text{th}}$ dimension of token $t_k^l$ and $\sigma(u) = \frac{1}{1+\text{exp}^{-u}}$ is the logistic sigmoid function. Above, $\beta$ and $\gamma$ are shifting and scaling parameters that adjust the embedding before applying the non-linearity. Note that these two scalar parameters are shared across all layers for all tokens. Only one entry of the embedding dimension $E$ is used for halting score calculation. Empirically, we observe that the simple choice of $e=0$ (the first dimension) performs well, while varying indices does not change the original performance, as we show later. As a result our halting mechanism does not introduce additional parameters or sub-network beyond the two scalar parameters $\beta$ and $\gamma$.

% \textbf{Training halting mechanism:} Here, we present our training formulation that trains our halting mechanism for adaptive early stopping. 
To track progress of halting probabilities across layers, we calculate a remainder for each token as:
\begin{equation}
   r_k = 1 - \sum_{l=1}^{N_k-1} h^l_k,
\end{equation}
that subsequent forms a halting probability as: 
% \AV{Should the probability be 0 or 1 for the first case?}:
\begin{equation}
    p^l_k = \begin{cases}
      0 & \text{if } \ l > N_k,\\
      r_k & \text{if } \ l = N_k, \\
      h^l_k & \text{if } \ l < N_k. \\
    \end{cases}  
\end{equation}
Given the range of $h$ and $r$, halting probability per token at each layer is always bounded $0\leq p^l_k \leq 1$. 
The overall ponder loss to encourage early stopping is formulated via auxiliary variable $r$ (reminder):
\begin{equation}
    \mathcal{L}_\text{ponder} := \frac{1}{K} \sum_{k=1}^{K} \rho_k = \frac{1}{K} \sum_{k=1}^{K} (N_k + r_k),
\end{equation}
% \AV{If I understand it correctly, ACT actually drops $N_K$ from its objective and only minimizes $\frac{1}{K} \sum_{k=1}^{K} r_k$ because $N_k$ is non-differentiable (See sec. 2.2 in ACT)If you have $N_k$ in your ponder loss, a question might be how you optimized $N_k$.}. \yin{yes}
where ponder loss  $\rho_k$ of each token is averaged. Vision transformers use a special class token $t_k$ to produce the classification prediction, we denote it as $t_c$ for future notations. This token similar to other input tokens is updated in all layers. We apply a mean-field formulation (halting-probability weighted average of previous states) to form the output token $t_o$ and the associated task loss as:
\begin{align} \label{eq:aggregation}
\mathcal{L}_\text{task} = \mathcal{C}(t_o), \ \text{where} \ t_o=\sum_{l=1}^{L} p_c^lt_c^l.
\end{align}
% \AV{Above, $\mathcal{L}_\text{task}$ and $\mathcal{C}(t_o)$ are not defined.}
% where  $\textbf{output}_k$ is the $k^{\text{th}}$ token among the total $K$ tokens in the $\textbf{output}$ tensor. \JK{this seems like a side-thought to put it here. Instead it deserves more attention, see comment earlier.}
Our vision transformer can then be trained by minimizing:
\begin{equation}
    \mathcal{L}_{\text{overall}} = \mathcal{L}_\text{task} + \alpha_{\text{p}}\mathcal{L}_\text{ponder},
\label{eqn:overall_loss_1}
\end{equation}
where $\alpha_{\text{p}}$ scales the pondering loss relative to the the main task loss. Algorithm~\ref{algo:main_algo} describes the overall computation flow, and Fig.~\ref{fig:main_algo} depicts the associated halting mechanism for visual explanation. 
% \AV{Danny: I updated the Algorithm to make sure that symbols match the main text. Please check them.} \AV{One thing I don't understand is that in Algorithm 1 we are computing the output features using meanfield for all tokens but in the text we say that we do that only for the output token. Shouldn't we talk about it in the text too?} \AV{Is $rho$ the same as $\mathcal{H}$? Please make sure that the symbols are consistent between figure, algorithm and text.}
At this stage, the objective function encourages an accuracy-efficiency trade-off when pondering different tokens at varying depths, enabling adaptive control.%
\begin{algorithm}[t]
\caption{Adaptive tokens in vision transformer without imposing extra parameters.}\label{algo:main_algo}
% \hspace*{\algorithmicindent} 
\textbf{Input:} tokenized input tensor \textbf{input} $\in \mathcal{R}^{K \times E}$, $K,E$ being token number and embedding dimension; $c$ is class-token index in $K$; $0<\epsilon<1$\\
% \hspace*{\algorithmicindent} 
\textbf{Output:} aggregated output tensor \textbf{out}, ponder loss $\rho$
    \begin{algorithmic}[1]
        \State $\mathbf{t}$ = \textbf{input} % $\mathbf{x}$ = \textbf{input}
        \State  $\mathbf{cumul}=\mathbf{0}$ \Comment{Cumulative halting score} 
        \State $\mathbf{R}=\mathbf{1}$ \Comment{Remainder value}
        \State \textbf{out} = $\mathbf{0}$ \Comment{Output of the network}
        \State $\boldsymbol{\rho} = \mathbf{0}$ \Comment{Token ponder loss vector}
        \State $\mathbf{m} = \mathbf{1}$ \Comment{Token mask $\mathbf{m} \in \mathcal{R}^{K}$}
        \For{$l=1 \ ... \ L$}
            \State $\mathbf{t}$ = $\mathcal{F}^l(\mathbf{t} \odot \mathbf{m} )$ % $\mathbf{x}$ = $\mathcal{F}^l(\mathbf{x} \odot \mathbf{m} )$
            \If{$l < L$} 
                \State $\mathbf{h} = \boldsymbol{\sigma}(\gamma \cdot \mathbf{t}_{:,0} + \beta)$ \hspace{-3mm} \Comment{$\mathbf{h} \in \mathcal{R}^{K}$}  % $\mathbf{h} = \boldsymbol{\sigma}(\mathbf{x}_{:,:,0} \cdot \gamma + \beta)$ \hspace{-3mm} \Comment{$\mathbf{h} \in \mathcal{R}^{K}$}
            \Else 
                \State \ $\mathbf{h}=\mathbf{1}$
            \EndIf
            \State $\mathbf{cumul} \mathrel{+}= \mathbf{h}$ 
            \State $\boldsymbol{\rho} \mathrel{+}= \mathbf{m}$ \Comment{Add one per remaining token}
            \For{$k = 1, ..., K$}
                \If{$\mathbf{cumul}_k<1-\epsilon$}
                    % \State \textbf{out}$_k \mathrel{+}= \mathbf{t}_{k,:} \times \mathbf{h}_k$
                    \State $\mathbf{R}_k \mathrel{-}= \mathbf{h}_k$
                \Else
                    % \State \textbf{out}$_k \mathrel{+}= \mathbf{t}_{k,:} \times \mathbf{R}_k$
                    \State $\boldsymbol{\rho}_k  \mathrel{+}= \mathbf{R}_k$
                \EndIf
                \iffalse
                \If{$k = c \ \text{and} \ \mathbf{cumul}_k<1-\epsilon$}
                    \State \textbf{out} $ \mathrel{+}= \mathbf{t}_{c,:} \times \mathbf{h}_k$
                    % \State $\mathbf{R}_k \mathrel{-}= \mathbf{h}_k$
                \Else
                    \State \textbf{out} $ \mathrel{+}= \mathbf{t}_{c,:} \times \mathbf{R}_k$
                    % \State $\boldsymbol{\rho}_k  \mathrel{+}= \mathbf{R}_k$
                \EndIf 
                \fi
            \EndFor
            \If{$\mathbf{cumul}_c<1-\epsilon$}
                \State \textbf{out} $ \mathrel{+}= \mathbf{t}_{c,:} \times \mathbf{h}_c$
                % \State $\mathbf{R}_k \mathrel{-}= \mathbf{h}_k$
            \Else
                \State \textbf{out} $ \mathrel{+}= \mathbf{t}_{c,:} \times \mathbf{R}_c$
                % \State $\boldsymbol{\rho}_k  \mathrel{+}= \mathbf{R}_k$
            \EndIf 
            \State $\mathbf{m} \leftarrow \mathbf{cumul}<1-\epsilon$ \Comment{Update mask}
        \EndFor
    \State \Return \textbf{out}, $\rho = \frac{\text{sum}(\boldsymbol{\rho})}{K}$ 
    \end{algorithmic}
\end{algorithm}%

One critical factor in Eqn.~\ref{eqn:overall_loss_1} is $\alpha_{\text{p}}$ that balances halting strength and network performance for the target application. A larger $\alpha_{\text{p}}$ value imposes a stronger penalty, and hence learns to halt tokens earlier. Despite efficacy towards computation reduction, prior work on adaptive computation~\cite{graves2016adaptive,figurnov2017spatially} have found that training can be sensitive to the choice of $\alpha_{\text{p}}$ and its value may not provide a fine-grain control over accuracy-efficiency trade-off. We empirically observe a similar behavior in vision transformers. 

% using one scaling coefficient as a mere control remains sensitive to hyper-parameter choice, as empirically observed in adaptive computation for CNNs, and we empirically observe in adaptive tokens for vision transformer. 

As a remedy, we introduce a distributional prior to regularize $h^l$ such that tokens are expected to exit at a target depth on average, however, we still allow per-image variations. In this case for infinite number of input images we expect the the depth of token to vary within the distributional prior. Similar prior distribution has been recently shown effective to stablize convergence during stochastic pondering~\cite{banino2021pondernet}. To this end, we define a halting score distribution:
\begin{equation}
    \mathcal{H} := [\frac{\sum_{k=1}^{K}h^1_k}{K}, \frac{\sum_{k=1}^{K}h^2_k}{K}, ..., \frac{\sum_{k=1}^{K}h^L_k}{K}],
\end{equation}
that averages expected halting score for all tokens across at each layer of network (\textit{i.e.}, $\mathcal{H} \in \mathcal{R}^L$). Using this as an estimate of how halting likelihoods distribute across layers, we regularize this distribution towards a pre-defined prior using KL divergence. We form the new distributional prior regularization term as:
\begin{equation}
    \mathcal{L}_{\text{distr.}} = \text{KL}(\mathcal{H} \ || \ \mathcal{H}^{\text{target}}),
\end{equation}
where KL refers to the Kullback-Leibler divergence, and $\mathcal{H}^{\text{target}}$ denotes a target halting score distribution with a guiding stopping layer. We use the probability density function of Gaussian distribution to define a bell-shaped distribution $\mathcal{H}^{\text{target}}$ in this paper, centered at the expected stopping depth $N^\text{target}$. %, and re-scaled to the summation of the first $N^\text{target}$ entries equals $1-\epsilon$. 
Intuitively, this weakly encourages the expected sum of halting score for each token to trigger exit condition at $N^\text{target}$. This offers enhanced control of expected remaining compute, as we show later in experiments. 

Our final loss function that trains the network parameters for adaptive token computation is formulated as:
\begin{equation}
    \mathcal{L}_{\text{overall}} = \mathcal{L}_\text{task} + \alpha_{\text{d}}\mathcal{L}_\text{distr.} +  \alpha_{\text{p}}\mathcal{L}_\text{ponder},
\label{eqn:overall_loss}
\end{equation}

\begin{equation}
    \mathcal{L}_{\text{overall}} = \mathcal{L}_\text{task} +  \alpha_{\text{p}}\mathcal{L}_\text{ponder}+ \alpha_{\text{d}}\mathcal{L}_\text{distr.} ,
\label{eqn:overall_loss}
\end{equation}

where $\alpha_{\text{d}}$ is a scalar coefficient that balances the distribution regularization against other loss terms.
% \AV{Instead of Gaussian, we could use Poisson distribution which is natural discrete version of the Gaussian distribution: \url{https://en.wikipedia.org/wiki/Poisson_distribution} }\yin{will try}
\section{Experiments}

\begin{figure*}[t]
\centering

\resizebox{\linewidth}{!}{
\begingroup
\renewcommand*{\arraystretch}{0.3}
\begin{tabular}{cccc}%original
\includegraphics[width=0.25\linewidth]{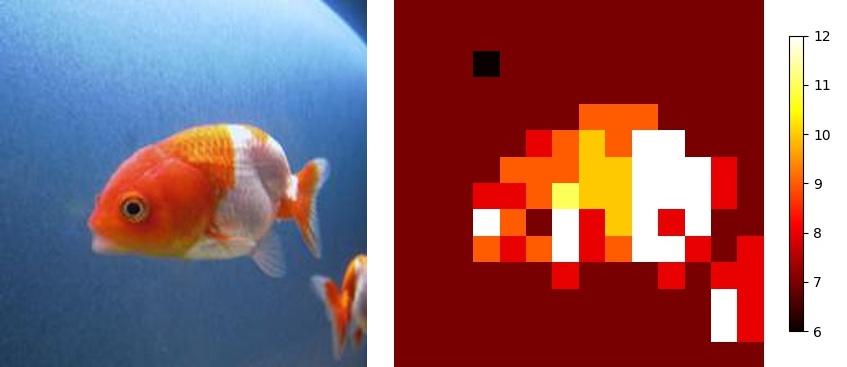} &
\includegraphics[width=0.25\linewidth]{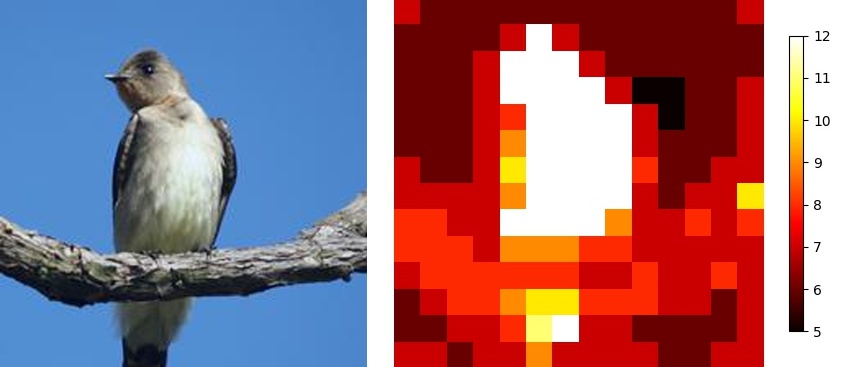} &
\includegraphics[width=0.25\linewidth]{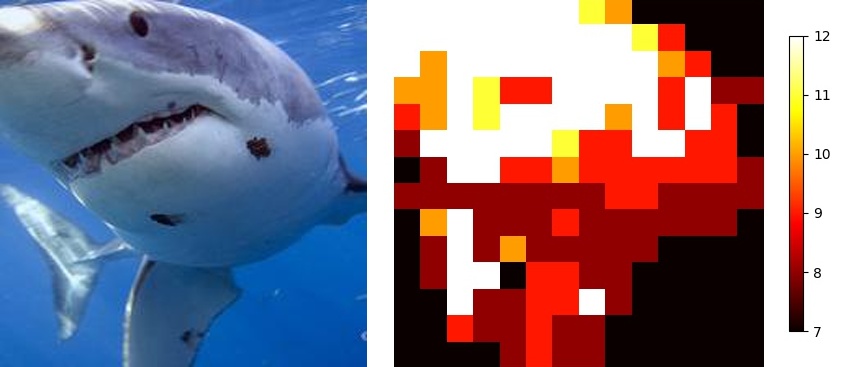} &
\includegraphics[width=0.25\linewidth]{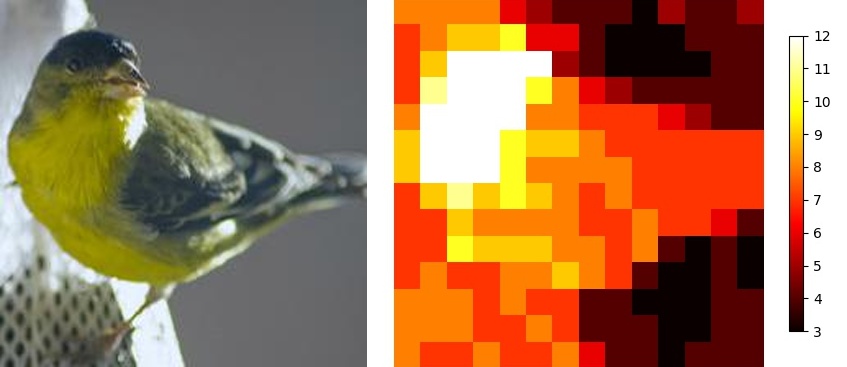} \\
\end{tabular}
\endgroup
}

\vspace{1mm}

\resizebox{\linewidth}{!}{
\begingroup
\renewcommand*{\arraystretch}{0.3}
\begin{tabular}{cccc}%original
\includegraphics[width=0.25\linewidth]{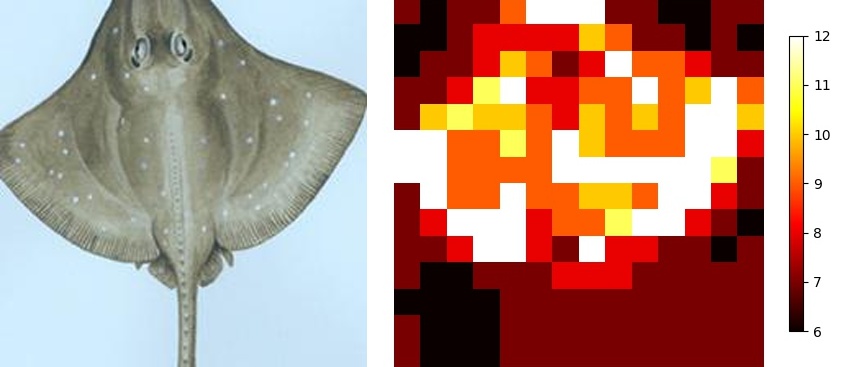} &
\includegraphics[width=0.25\linewidth]{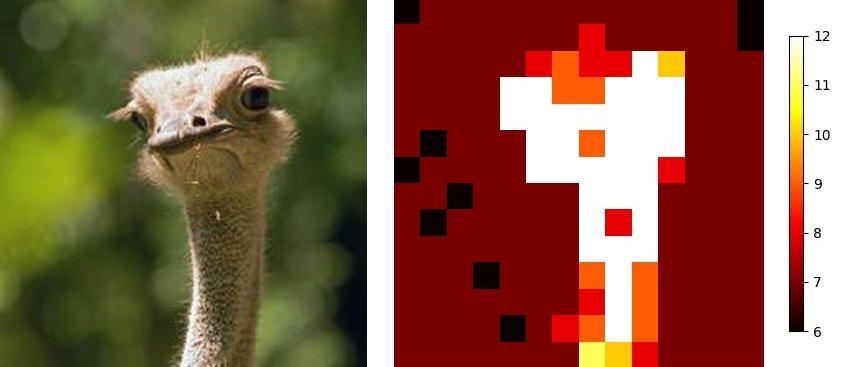} &
\includegraphics[width=0.25\linewidth]{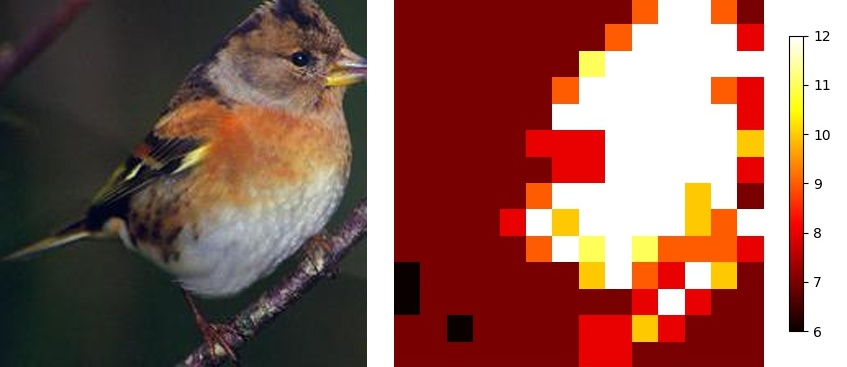} &
\includegraphics[width=0.25\linewidth]{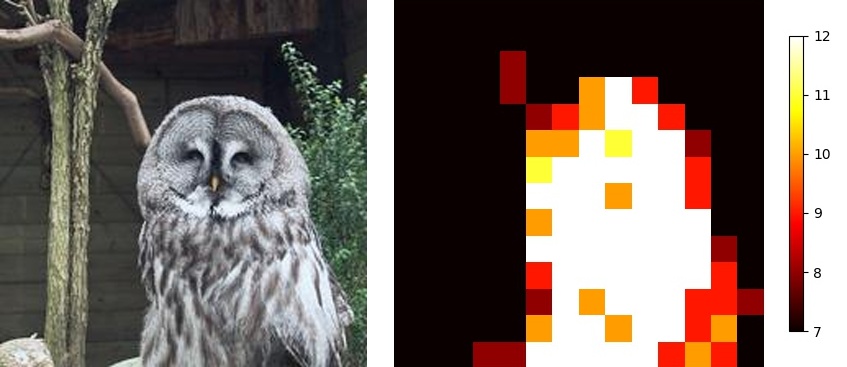}  \\
\end{tabular}
\endgroup
}

\vspace{1mm}

\resizebox{\linewidth}{!}{
\begingroup
\renewcommand*{\arraystretch}{0.3}
\begin{tabular}{cccc}%original
\includegraphics[width=0.25\linewidth]{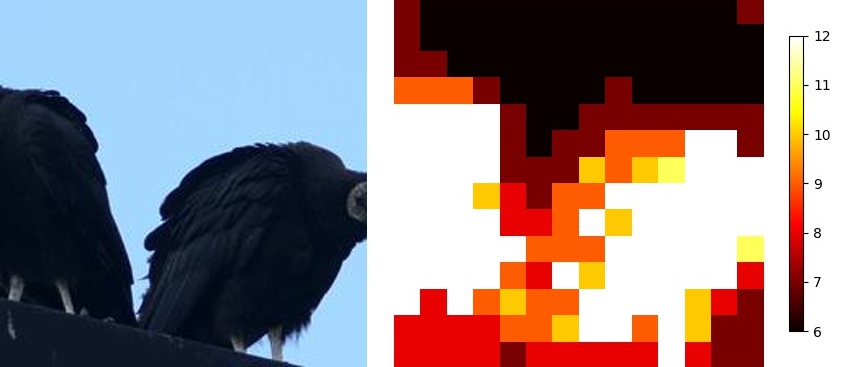} &
\includegraphics[width=0.25\linewidth]{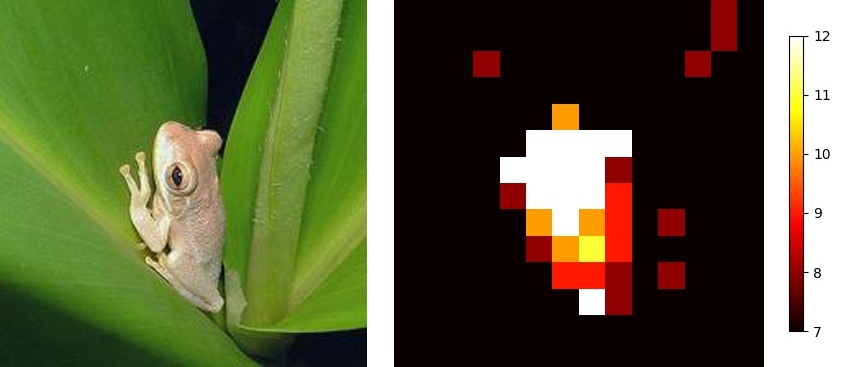} &
\includegraphics[width=0.25\linewidth]{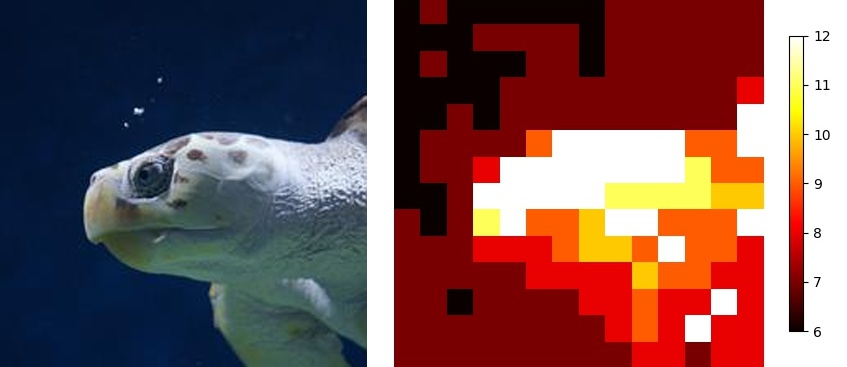} &
\includegraphics[width=0.25\linewidth]{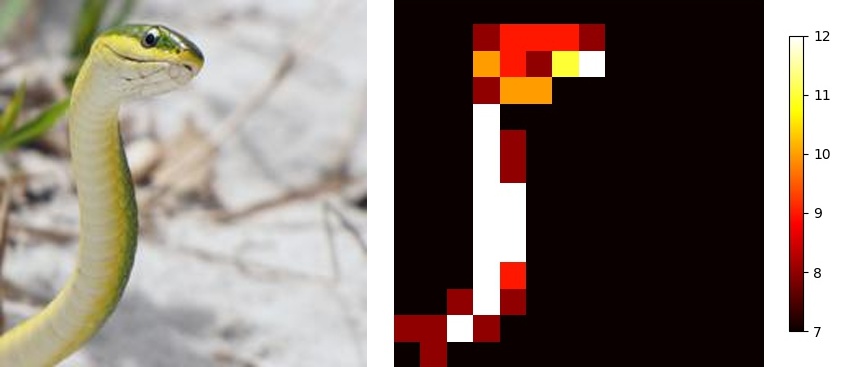} \\
\end{tabular}
\endgroup
}

\vspace{2mm}

\resizebox{\linewidth}{!}{
\begingroup
\renewcommand*{\arraystretch}{0.3}
\begin{tabular}{cccc}%original
\includegraphics[width=0.25\linewidth]{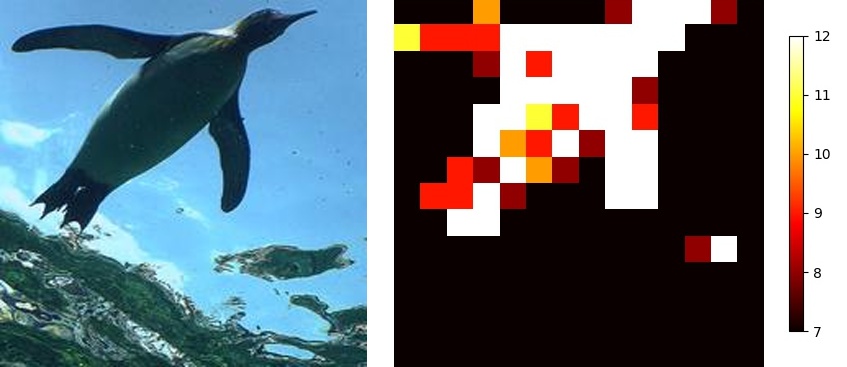} & 
\includegraphics[width=0.25\linewidth]{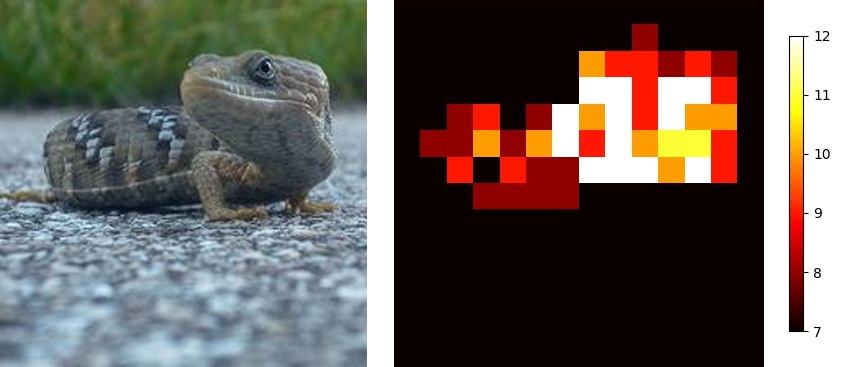} &
\includegraphics[width=0.25\linewidth]{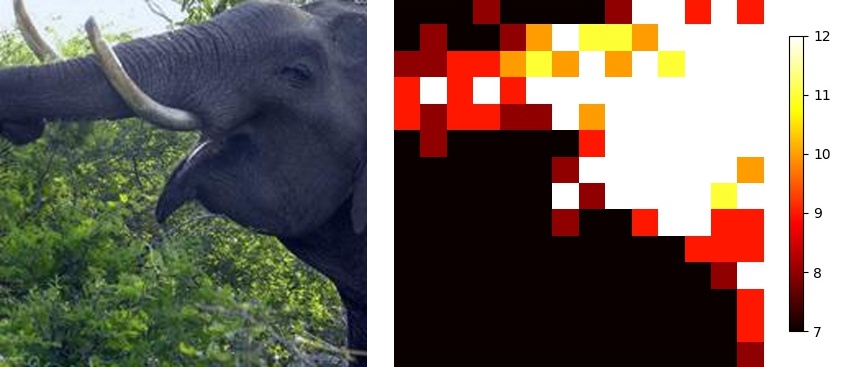} &
\includegraphics[width=0.25\linewidth]{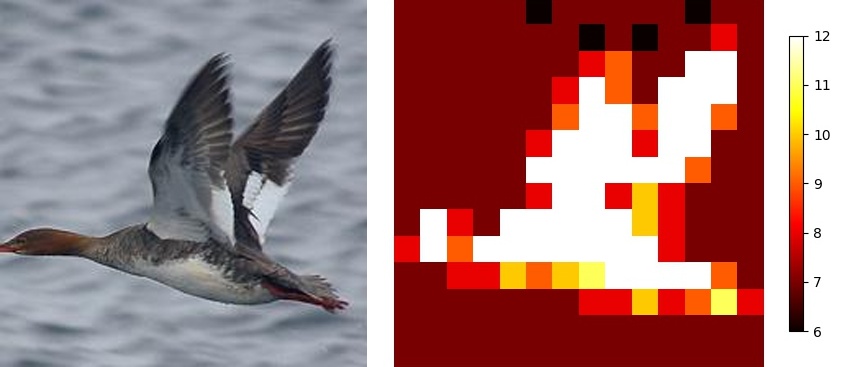}  \\
\end{tabular}
\endgroup
}

\vspace{2mm}

\resizebox{\linewidth}{!}{
\begingroup
\renewcommand*{\arraystretch}{0.3}
\begin{tabular}{cccc}%original
\includegraphics[width=0.25\linewidth]{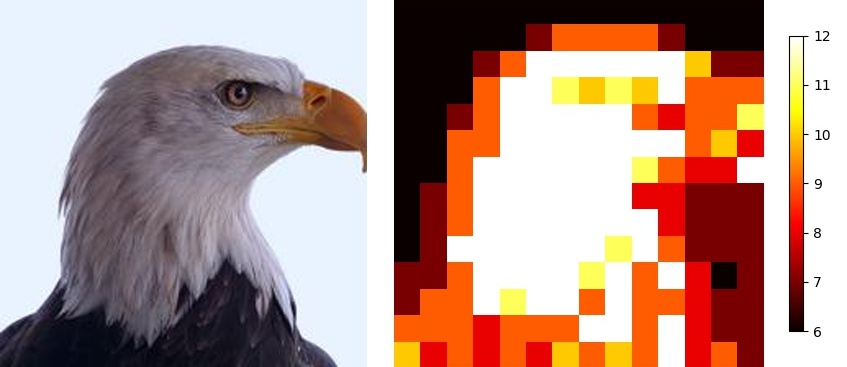} &
\includegraphics[width=0.25\linewidth]{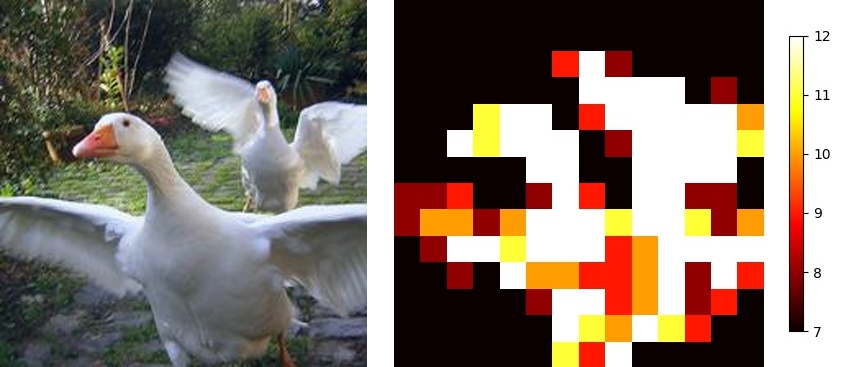} &
\includegraphics[width=0.25\linewidth]{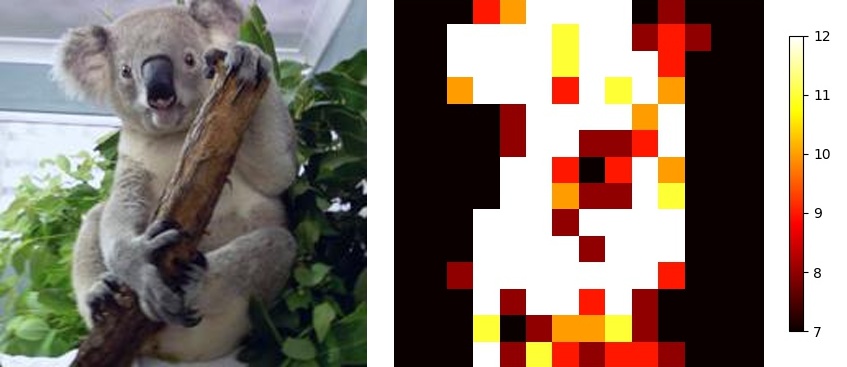} &
\includegraphics[width=0.25\linewidth]{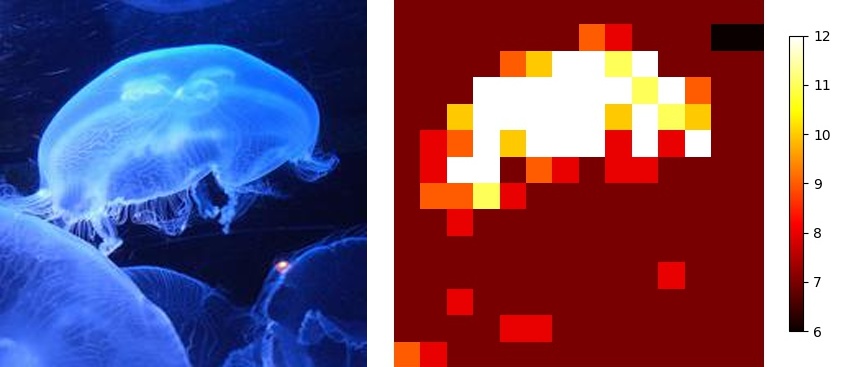} \\
\end{tabular}
\endgroup
}

\vspace{2mm}

\resizebox{\linewidth}{!}{
\begingroup
\renewcommand*{\arraystretch}{0.3}
\begin{tabular}{cccc}%original
\includegraphics[width=0.25\linewidth]{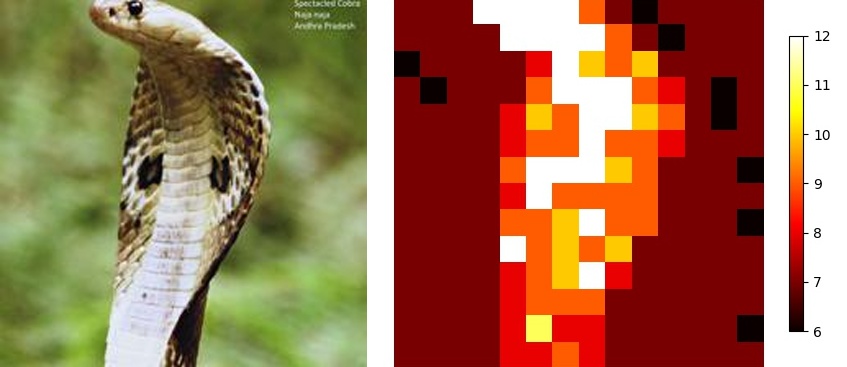} &
\includegraphics[width=0.25\linewidth]{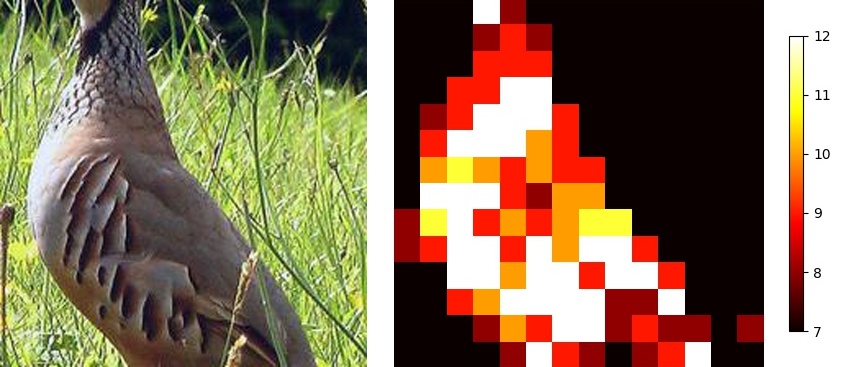} &
\includegraphics[width=0.25\linewidth]{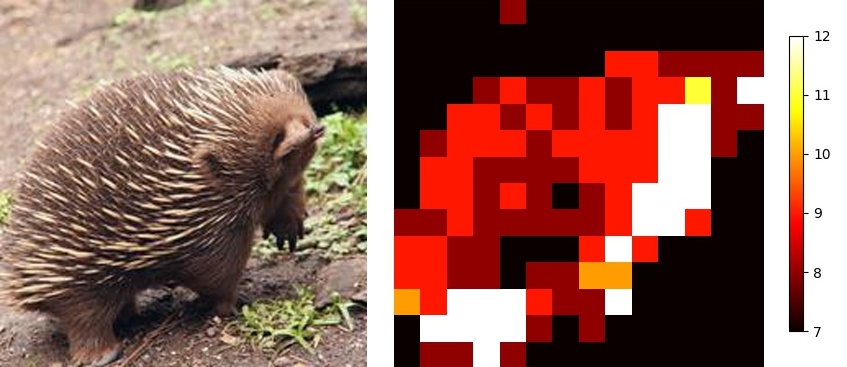} &
\includegraphics[width=0.25\linewidth]{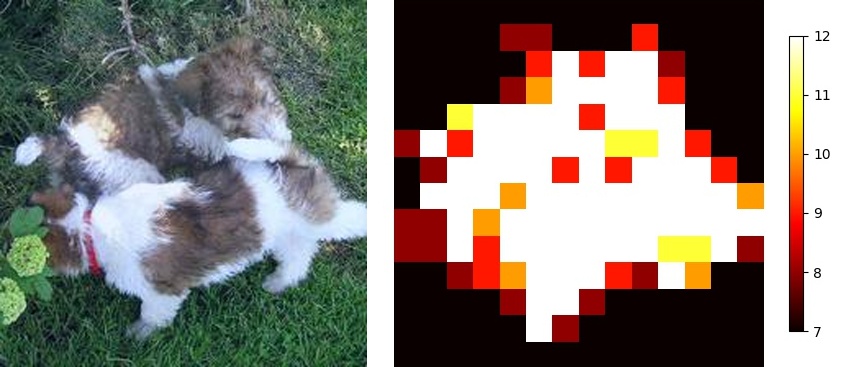} \\
\end{tabular}
\endgroup
}

\vspace{2mm}

\resizebox{\linewidth}{!}{
\begingroup
\renewcommand*{\arraystretch}{0.3}
\begin{tabular}{cccc}%original
\includegraphics[width=0.25\linewidth]{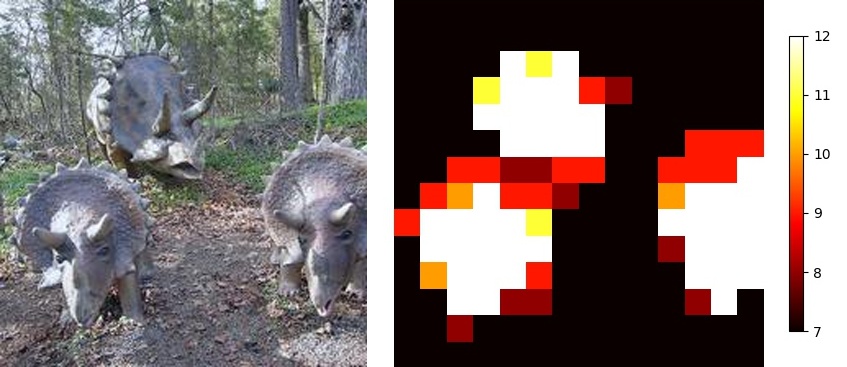} &
\includegraphics[width=0.25\linewidth]{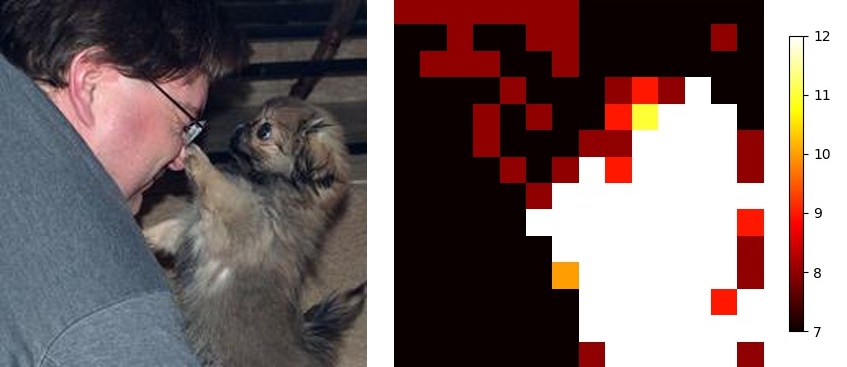} &
\includegraphics[width=0.25\linewidth]{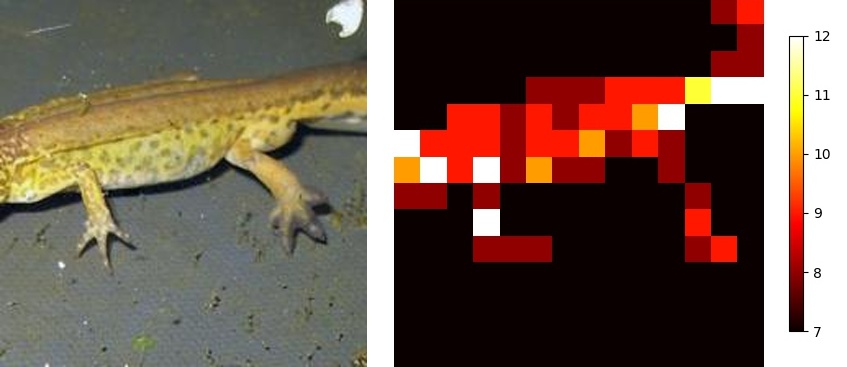} &
\includegraphics[width=0.25\linewidth]{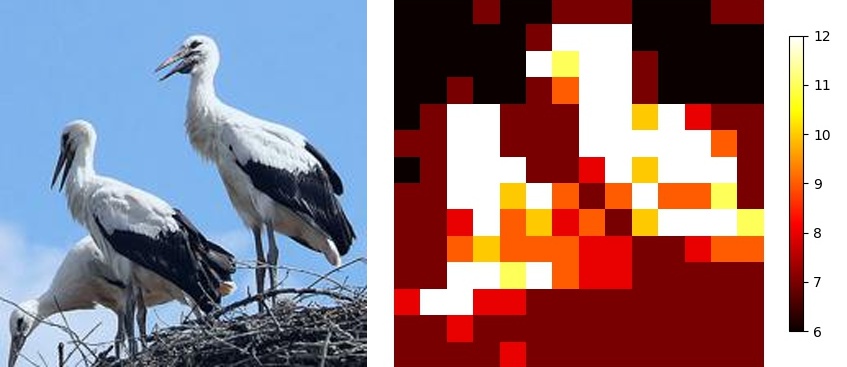} \\
\end{tabular}
\endgroup
}

\vspace{1mm}

\caption{Original image (left) and the dynamic token depth (right) of A-ViT-T on the ImageNet-1K validation set. \textbf{Distribution of token computation highly aligns with visual features.} Tokens associated with informative regions are adaptively processed deeper, robust to repeating objects with complex backgrounds. Best viewed in color.}
\label{fig:all_images}
\end{figure*}

We evaluate our method for the classification task on the large-scale $1000$-class ImageNet ILSVRC 2012 dataset~\cite{deng2009imagenet} at the $224\times224$ pixel resolution. We first analyze the performance of adaptive tokens, both qualitatively and quantitatively. Then, we show the benefits of the proposed method over prior art, followed by a demonstration of direct throughput improvements of vision transformers on legacy hardware. Finally, we evaluate the different components of our proposed approach to validate our design choices.

\noindent\textbf{Implementation details.}
We base A-ViT on the data-efficient vision transformer architecture (DeiT)~\cite{touvron2021training} that includes $12$ layers in total. Based on original training recipe\footnote{Based on official repository at \url{https://github.com/facebookresearch/DeiT}.}, we train all models on only ImageNet1K dataset without auxiliary images. We use the default $16\times16$ patch resolution. For all experiments in this section, we use Adam for optimization (learning rate $1.5\cdot 10^{-3}$) with cosine learning rate decay. For regularization constants we utilize $\alpha_{\text{d}}=0.1, \alpha_{\text{p}}=5\cdot10^{-4}$ to scale loss terms. We use $\gamma=5, \beta= -10$ for sigmoid control gates $H(\cdot)$, shared across all layers. We use the embedding value at index $e=0$ to represent the halting probability ($H(\cdot)$) for tokens. Starting from publicly available pretrained checkpoints, we fine-tune DeiT-T/S variant models for $100$ epochs, respectively, to learn adaptive tokens without distillation. We denote the associated adaptive versions as A-ViT-T/S respectively. In what follows, we mainly use the A-ViT-T for ablations and analysis before showing efficiency improvements for both variants afterwards. We find that mixup is not compatible with adaptive inference,
% \JK{the following is unclear}\yin{modified} 
and we focus on classification without auxiliary distillation token -- we remove both from finetuning. Applying our finetuning
% \JK{"to enable halting scores"}\yin{this is our finetuning baseline.}
on the full DeiT-S and DeiT-T results in a top-1 accuracy of $78.9\%$ and $71.3\%$, respectively. For training runs we use $8$ NVIDIA V100 GPUs and automatic-mixed precision (AMP)~\cite{micikevicius2017mixed} acceleration. 
% \JK{say how this compares to non-finetuned?}\yin{there is an accuracy drop here ~0.9 so we did not explicitly report, and will try to improve after submission.}

\subsection{Analysis}

\noindent
\textbf{Qualitative results.} Fig.~\ref{fig:all_images} visualizes the tokens' depth that is adaptively controlled during inference with our A-ViT-T over the ImageNet1K validation set. Remarkably, we observe that our adaptive token halting enables longer processing for highly discriminative and salient regions, often associated with the target class. Also, we observe a highly effective halting of relatively irrelevant tokens and their associated computations. For example, our approach on animal classes retains the eyes, textures, and colors from the target object and analyze them in full depth, while using fewer layers to process the background (\textit{e.g.}, the sky around the bird, and ocean around sea animals). Note that even background tokens marked as not important still actively participate in classification during initial layers. In addition, we also observe the inspiring fact that adaptive tokens can easily (i) keep track of repeating target objects, as shown in the first image of the last row in Fig.~\ref{fig:all_images}, and (ii) even shield irrelevant objects completely (see second image of last row).

\begin{figure}[t]
\begingroup
\vspace{-3mm}
\begin{tabular}{cc}
\centering
\hspace{-5mm} \includegraphics[width=0.5\linewidth,trim={0 0 0 0}]{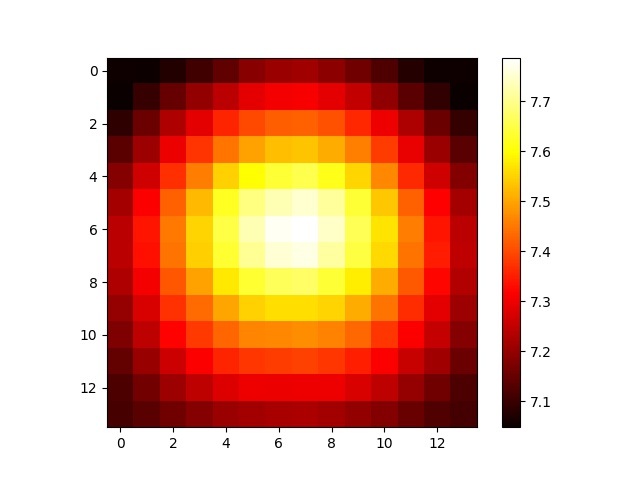} &
\includegraphics[width=0.5\linewidth,trim={0pt 0pt 0pt 30pt},clip]{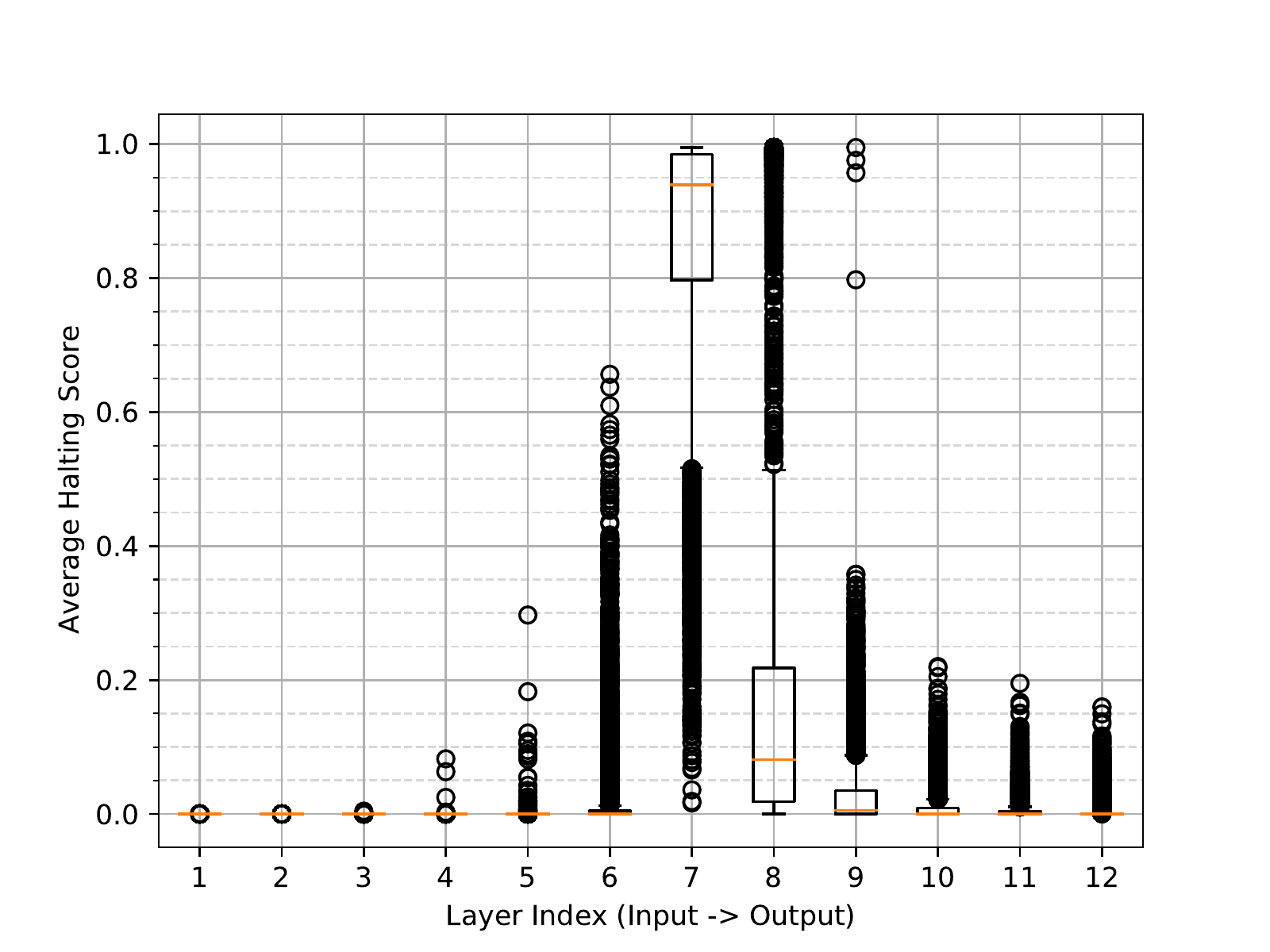} \\
\hspace{-2.5mm} \small{(a)} & \small{(b)} 
\end{tabular}
\endgroup
\caption{(a) Average depth of tokens per image patch position for A-ViT-T on ImageNet-1K validation set. (b) Halting score distribution across the transformer blocks. Each point associated with one randomly sampled image, denoting average token score at that layer. }
\label{fig:average_depth}
\end{figure}

\noindent
\textbf{Token depth distribution.} Given a complete distinct token distribution per image, we next analyze the dataset-level token importance distributions for additional insights. Fig.~\ref{fig:average_depth}~(a) depicts the average depth of the learnt tokens over the validation set. It demonstrates a 2D Gaussian-like distribution that is centered at the image center. This is consistent with the fact that most ImageNet samples are centered, intuitively aligning with the image distribution. As a result, more compute is allocated on-the-fly to center areas, and computational cost on the sides is reduced. 
% \begin{figure}[t]
% \centering
% \includegraphics[width=0.6\linewidth,trim={0 0.3cm 0 0}]{fig/tiny_token_depths/accum_token.jpg}
% \caption{Average depth of tokens per image patch position for A-ViT-T on ImageNet-1K validation set.}
% \vspace{-2mm}
% \label{fig:average_depth}
% \end{figure}

\noindent\textbf{Halting score distribution.} 
To further evaluate the halting behavior across transformer layers, we plot the average layer-wise halting score distribution over $12$ layers. Fig.~\ref{fig:average_depth}~(b) shows box plots of halting scores averaged over all tokens per layer per image. The analysis is performed on $5$K randomly sampled validation images. As expected, the halting score gradually increases at initial stages, peaks and then decreases for deeper layers. 
% \PM{Danny to check -$>$} Hating scores also vary greatly across samples as shown in the wide span of dots at each layer, showing the high efficacy to adjust halting to varying sample complexities.   
% \begin{figure}[t]
% \centering
% % \includegraphics[width=0.65\linewidth]{fig/halting_score.png}
% \includegraphics[width=0.6\linewidth,trim={0pt 0pt 0pt 30pt},clip]{fig/box_plot.pdf}
% \caption{Halting score distribution across the transformer blocks. Each point associated with one randomly sampled image, denoting average token score at that layer. The figure is based on $5$K randomly sampled validation images.}
% \label{fig:halting_score}
% \end{figure}

\noindent
\textbf{Sharp-halting baseline.} To further compare with static models of the same depth for performance gauging, we also train a DeiT-T with $8$ layers as a sharp-halting baseline. We observe that our A-ViT-T outperforms this new baseline by $+1.4\%$ top-1 accuracy at a similar throughput. Although our adaptive regime is on average similarly shallow, it still inherits the expressivity of the original deeper network, as we observe that informative tokens are processed by deeper layers (\textit{e.g.}, until $12^{\text{th}}$ layer as in Fig.~\ref{fig:all_images}).

\begin{figure}[t!]
\centering
\begingroup
\begin{tabular}{c|c}
\centering
\includegraphics[width=0.3\linewidth]{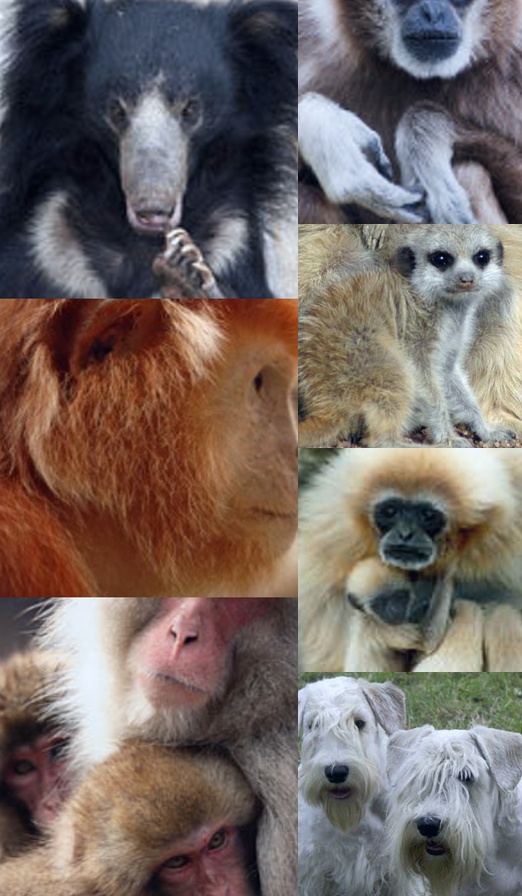} &
\includegraphics[width=0.3\linewidth]{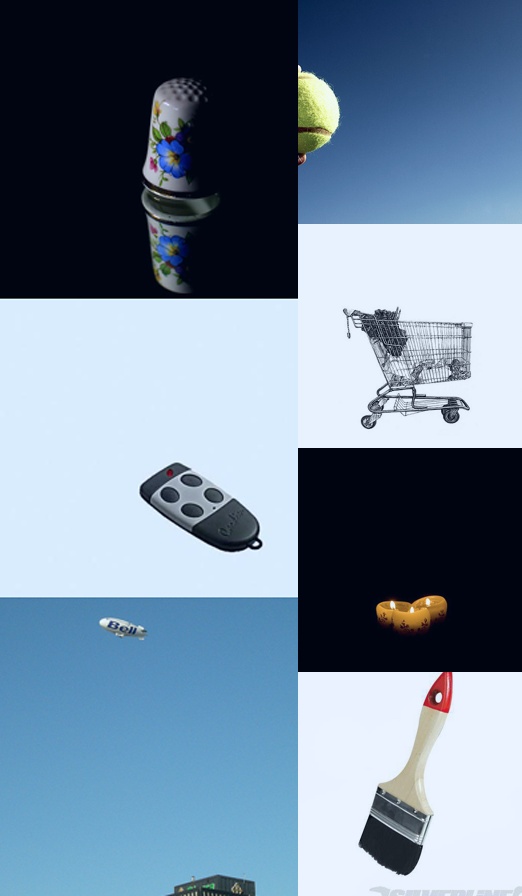} \\
\small{Hard samples.} & \small{Easy samples.} 
\end{tabular}
\endgroup
\caption{Visual comparison of hard and easy samples from the ImageNet-1K validation set determined by average token depth. Note that all images above be \textit{correctly classified} -- only difference is that hard samples require more depths for tokens to process their semantic information. Tokens in the left images exit approximately~$5$ layers later compared to the right images.}
\label{fig:sample_hardness}
\end{figure}

\noindent\textbf{Easy and hard samples.} We can analyse the difficulty of an image for the network by looking at the averaged depth of the adaptive tokens per image. Therefore, in Fig.~\ref{fig:sample_hardness}, we depict hard and easy samples in terms of the required computation. Note, all samples in the figure are correctly classified, and only differ by the averaged token depth. We can observe that images with homogeneous background are relatively easy for classification, and A-ViT processes them much faster than hard samples. Hard samples represent images with informative visual features distributed over the entire image, and hence incur more computation.

\noindent\textbf{Class-wise sensitivity.} Given an adaptive inference paradigm, we analyze the change in classification accuracy for various classes with respect to the full model.
% instead of the conventionally static one, we try to analyze the shift's specific impact on different class categories. 
In particular, we compute class-wise validation accuracy changes before and after applying adaptive inference. We summarize both qualitative and quantitative results in Table~\ref{tab:per_class_analysis}. We observe that originally very confident or uncertain samples are not affected by adaptive inference. Adaptive inference improves accuracy of the visually dominant classes such as individual furniture and animals.
% Static inference is more adept at capturing interactions between features that embed in correlated contexts spanning across entire image as in household items, while dynamic inference excels in classifying 
% \yin{any recommendation? the last sentence is vague.}
\begin{table}[t]
\centering

\resizebox{\linewidth}{!}{
\begin{tabular}{cccc}
\toprule
\multirow{2}{*}{\textbf{Rank}} & \multicolumn{3}{c}{\textbf{Class-wise Sensitivity to Adaptive Inference} $_{\text{static acc.} \rightarrow \text{adaptive acc.}}$} \\
\cmidrule{2-4} 
& Favoring (acc. incr.) & Sensitive (acc. drop) & Stable \\
\midrule
$1$ & throne $_{56\rightarrow74\%}$ & muzzle $_{58\rightarrow38\%}$ & yellow lady-slipper $_{100\rightarrow100\%}$ \\
$2$ & lakeland terrier $_{64\rightarrow78\%}$  &  sewing machine $_{80\rightarrow62\%}$ & leonberg $_{100\rightarrow100\%}$ \\
$3$ & cogi $_{60\rightarrow74\%}$ & vaccume $_{37\rightarrow28\%}$ & proboscis monkey $_{100\rightarrow100\%}$\\
\cmidrule{4-4} 
$4$ & african elephant $_{54\rightarrow68\%}$ & flute  $_{38\rightarrow20\%}$ & velvet $_{10\rightarrow10\%}$ \\
$5$ & soft-coated wheaten terrier $_{68\rightarrow82\%}$ & shovel $_{64\rightarrow46\%}$ & laptop $_{14\rightarrow14\%}$ \\
\bottomrule
\end{tabular}}

\resizebox{0.99\linewidth}{!}{
\begingroup
\begin{tabular}{ccc|ccc}
% grad only
\centering
\includegraphics[width=0.2\linewidth]{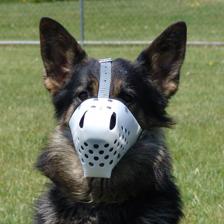} &
\includegraphics[width=0.2\linewidth]{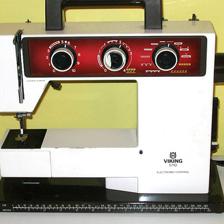} & 
\includegraphics[width=0.2\linewidth]{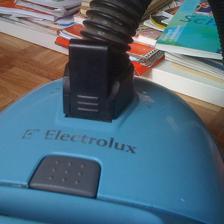}  &
\includegraphics[width=0.2\linewidth]{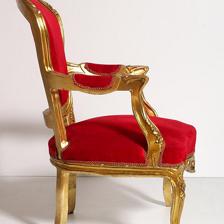} &
\includegraphics[width=0.2\linewidth]{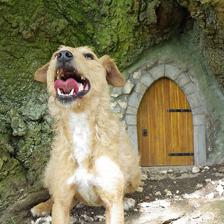} &
\includegraphics[width=0.2\linewidth]{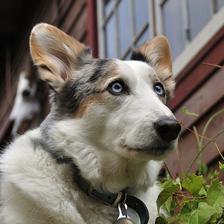}
\\ 
% \includegraphics[width=0.1\linewidth]{fig/sensitive_analysis/flute.jpg} & 
% \includegraphics[width=0.1\linewidth]{fig/sensitive_analysis/shovel.jpg}
% $\mathcal{N}(0, \mathcal{I})$ & 
\small{muzzle} &
\small{sewing machine} & 
\small{vacuum}  &
\small{throne} &
\small{terrier} &
\small{cogi}
\\
 \multicolumn{3}{c}{fixed  {\color{green}{\cmark}} adaptive {\color{red}{\xmark}}}
 &
 \multicolumn{3}{c}{ fixed {\color{red}{\xmark}} adaptive {\color{green}{\checkmark}}} 
\end{tabular}
\endgroup
}\vspace{-3mm}
\caption{Ranking of stable and sensitive classes to adaptive computation in A-ViT compared to fixed computation graph that executes the full model for inference. Sample images included for top three classes that favor or remain sensitive to adaptive computation.
% \AV{What are the images in the bottom representing? I think the caption may need a bit more description of what's being shown.}
}
\vspace{-3mm}
\label{tab:per_class_analysis}
\end{table}

% \begin{table}[!t]
% \centering
% \resizebox{\linewidth}{!}{
% \begin{tabular}{cccc}
% \toprule
% \multirow{2}{*}{\textbf{Rank}} & \multicolumn{3}{c}{\textbf{Class-wise Response to Dynamic Tokens}} \\
% \cmidrule{2-4} 
% & Favoring (acc. incr.) & Sensitive (acc. drops) & Stable \\
% \midrule
% $1$ & Throne $_{56\rightarrow74\%}$ & muzzle $_{58\rightarrow38\%}$ & yellow lady-slipper $_{100\rightarrow100\%}$ \\
% $2$ & Lakeland terrier $_{64\rightarrow78\%}$  &  sewing machine $_{80\rightarrow62\%}$ & leonberg $_{100\rightarrow100\%}$ \\
% $3$ & Cogi $_{60\rightarrow74\%}$ & horned viper $_{64\rightarrow46\%}$ & proboscis monkey $_{100\rightarrow100\%}$\\
% \cmidrule{4-4} 
% $4$ & African elephant $_{54\rightarrow68\%}$ & flute  $_{38\rightarrow20\%}$ & velvet $_{10\rightarrow10\%}$ \\
% $5$ & Soft-coated wheaten terrier $_{68\rightarrow82\%}$ & shovel $_{64\rightarrow46\%}$ & laptop $_{14\rightarrow14\%}$ \\

% \bottomrule
% \end{tabular}}
% \caption{Ranking of stable and sensitive classes to dynamic token inference. \yin{I can include images here too. Open to suggestions.}}
% \label{tab:per_class_analysis}
% \end{table}

% \noindent\textbf{Robustness to encoding position.} Embedding offers extra robustness to use as free ACT score source. We will include results here per accuracy loss after 10 random runs with stds, etc.

% \noindent\textbf{Semantic analysis.} We want to check whether the uncovered spaces co-alligns with segmentation network. 

\subsection{Comparison to Prior Art}

Next, we compare our method with previous work that study adaptive computation. For comprehensiveness, we systematically compare with five state-of-the-art halting mechanisms, covering both vision and NLP methods that tackle the dynamic inference problem from different perspectives: (i) adaptive computation time~\cite{graves2016adaptive} as ACT reference applied on halting entire layers, 
% \AV{(which corresponds to training the model with Eq.~\ref{eqn:overall_loss_1})}, \yin{not all as they are in fact layer-wise and eqn. is token wise}
(ii) confidence-based halting~\cite{weijie2020fastbert} that gauges on logits, (iii) similarity-based halting~\cite{elbayad2019depth} that oversees layer-wise similarity, (iv) pondering-based halting~\cite{banino2021pondernet} that exits based on stochastic halting-probabilities, and (v) the very recent DynamicViT~\cite{rao2021dynamicvit} that learns halting decisions via Gumble-softmax relaxation. Details in appendix.
% \begin{comment}
% \begin{itemize}
%     \item \textit{Adaptive computation time}~\cite{graves2016adaptive}. We start with comparing to conventional ACT that halts the computation of the network at a layer-level based on class token.  
%     \item \textit{Confidence-based halting}~\cite{weijie2020fastbert}. In addition, we gauge the performance of adaptive halting via stopping inference once the class-token confidence surpasses a pre-defined threshold, as studied in literature. 
%     \item \textit{Similarity-based halting}~\cite{elbayad2019depth}. Further, we compare to the recent stream of work that stops inference once the classification logits stabilize between adjacent layers. 
%     \item \textit{Pondering-based halting}~\cite{banino2021pondernet}.
%     We also compare to PonderNet~\cite{banino2021pondernet} that learns the halting probability via regularization, and adaptively halts compute through stochastic exit.
%     \item \textit{Gumbel-based halting}~\cite{rao2021dynamicvit}.
%     Finally we compare with the very recent DynamicViT~\footnote{Based on official paper repository at \url{https://github.com/raoyongming/DynamicViT}.}~\cite{rao2021dynamicvit} that relaxes token-level halting binary decisions into gumble-softmax distribution, and then learns halting mechanism jointly with network weights.
% \end{itemize}
% \end{comment}

\begin{figure}[t]
\centering

\resizebox{0.95\linewidth}{!}{
\begingroup
\renewcommand*{\arraystretch}{0.3}
\begin{tabular}{ccc}
%original
\includegraphics[width=0.33\linewidth,clip,trim=5px 0 0 4px]{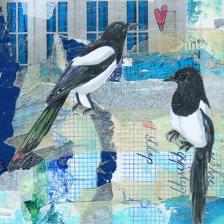} &
%ADI
\includegraphics[width=0.33\linewidth]{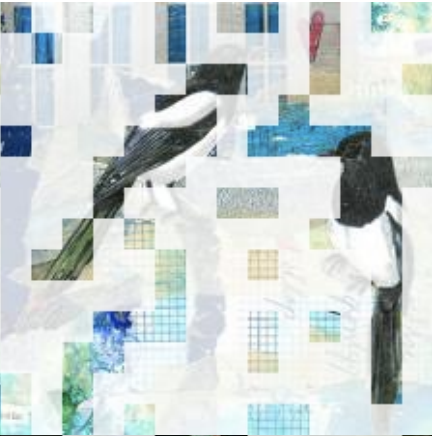} &
%deepdream
\includegraphics[width=0.33\linewidth,clip,trim=5px 0 0 4px]{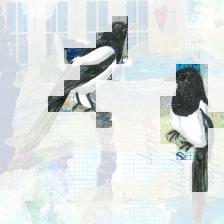}
\\
Original & DynamicViT (Rao \textit{et al.}~\cite{rao2021dynamicvit}) & \textbf{Ours}
\\
\end{tabular}
\endgroup
}

\caption{Visual comparison compared to prior art on token distribution for a sample taken from the public repository of DynamicViT by Rao \textit{et al.}~\cite{rao2021dynamicvit}. Only shaded (non-white) tokens are processed by all $12$ layers. Our method better captures the semantics of the target class, drops more tokens, and saves more computation.  
%Results of prior work based on the public repositories from the authors at~\cite{zgeiping_github,zzhu_github}. 
}
\label{fig:against_dynamic_vit_qual}
\end{figure}

\noindent\textbf{Performance comparison.} We compare our results in Table~\ref{tab:against_sota} and demonstrate simultaneous performance improvements over prior art in having smaller averaged depth, smaller number of FLOPs and better classification accuracy. Notably our method involves no extra parameters, while cutting down FLOPs by $39\%$ with only a minor loss of accuracy. 
To further visualize improvements over the state-of-the-art DynamicViT~\cite{rao2021dynamicvit}, we include Fig.~\ref{fig:against_dynamic_vit_qual} as a qualitative comparison of token depth for an official sample presented in the work. As noticed, A-ViT more effectively captures the important regions associated with the target objects, ignores the background tokens, and improves efficiency. 

Note that both DynamicViT and A-ViT investigate adaptive tokens but from two different angles. DynamicViT utilizes Gumbel-Softmax to learn halting and incorporates a control for computation via a multi-stage token keeping ratio; it provides stronger guarantees on the latency by simply setting the ratio. A-ViT on the other hand takes a complete probabilistic approach to learn halting via ACT. This enables it to freely adjust computation, and hence capture enhanced semantic and improve accuracy, however requires a distributional prior and has a less intuitive hyper-parameter.

\begin{table}[!t]
\centering
\resizebox{\linewidth}{!}{
\begin{tabular}{lcrrr}
\toprule
\multirow{2}{*}{\textbf{Method}} & \multicolumn{3}{c}{\textbf{Efficiency}} & \multirow{2}{*}{\textbf{Top-1 Acc.} $\uparrow$} \\
\cmidrule{2-4}
& Params. free & Avg. depth $\downarrow$ & FLOPs $\downarrow$  & \\
\midrule
Baseline~\cite{touvron2021training}& - & $12.00$ & $1.3$G & $71.3$ \\
\midrule
ACT~\cite{graves2016adaptive}& \xmark & $10.01$ & $1.0$G &  $71.0$ \\
Confidence threshold~\cite{weijie2020fastbert} & \cmark & $10.63$ & $1.1$G &  $65.8$ \\
Similarity gauging~\cite{elbayad2019depth} & \cmark & $10.68$ & $1.1$G & $69.4$\\
PonderNet~\cite{banino2021pondernet} & \cmark & $9.74$ & $1.0$G & $66.2$  \\
DynamicViT~\cite{rao2021dynamicvit} 
& \xmark & $7.62$ & $0.9$G  & $70.9$ \\
% \midrule
% \textbf{Ours} & \cmark & $\mathbf{8.84}$ & $\mathbf{0.9}$G & $\mathbf{72.2}$ \\
\rowcolor{lgreen}
\textbf{Ours} & \cmark & $\mathbf{7.23}$ & $\mathbf{0.8}$G & $\mathbf{71.0}$ \\
\bottomrule
\end{tabular}}
\caption{Comparison with prior art that studies dynamic inference halting mechanisms for transformers. Avg. depth specifies the mean depths of the tokens over the entire validation set.}
\vspace{-2mm}
\label{tab:against_sota}
\end{table}

\noindent\textbf{Hardware speedup.} In Table~\ref{tab:latency}, we compare speedup on off-the-shelf GPUs. See appendix for measurement details. In contrast to spatial ACT in CNNs that require extra computation flow and kernel re-writing~\cite{figurnov2017spatially}, A-ViT enables speedups out of the box in vision transformers. 
% , specifically TITAN RTX as inference platform. Throughput is measured with PyTorch at batch size $64$. 
With only $0.3\%$ in accuracy drop, our method directly improves the throughputs of DeiT small and tiny variants by $38\%$ and $62\%$ without requiring hardware/library modification.

\subsection{Ablations}
Here, we perform ablations studies to evaluate each component in our method and validate their contributions.

\noindent\textbf{Token-level ACT via $\mathcal{L}_{\text{ponder}}$.} One noticeable distinction of this work from conventional ACT~\cite{graves2016adaptive} is a full exploration of spatial redundancy in image patches, and hence their tokens. Comparing the first and last row in Table~\ref{tab:against_sota}, we observe that our fine-grained pondering reduces token depths by roughly $3$ layers, and results in $25\%$ more FLOP reductions compared to the conventional ACT. 

\begin{figure}[t]
\centering
\includegraphics[width=\linewidth]{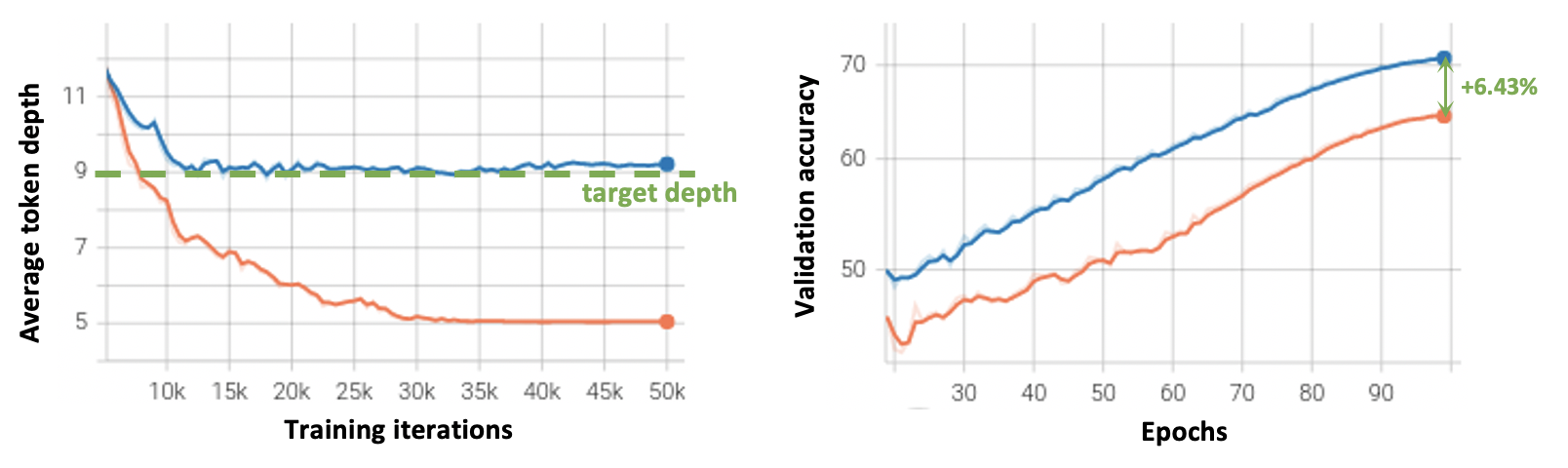}
\caption{Training curves with (blue) and without (yellow) distributional priors towards a target depth of $9$ layers. Both lines share the exact same training hyper-parameter set with the only difference in including the distributional prior guidance. As opposed to $\alpha_\text{p}$ 
% \AV{$\mathcal{L}_\text{ponder.}$?} \yin{in fact alpha as it's key that results in unstableness}
that over-penalizes the networks, $\mathcal{L}_\text{distr.}$ 
% \AV{$\mathcal{L}_\text{distr.}$?} 
guides a very fast convergence towards the target depth and yields a $6.4\%$ accuracy gain.
}
\label{fig:distr_prior}
\end{figure}

\noindent\textbf{Distributional prior via $\mathcal{L}_{\text{distr.}}$.} Incorporating the distributional prior allows us to better guide the expected token depth towards a target average depth, as seen in Fig.~\ref{fig:distr_prior}. As opposed to $\alpha_\text{p}$ that indirectly gauges on the remaining efficiency and usually suffers from over-/under-penalization, our distributional prior guides a quick convergence to a target depth level, and hence improves final accuracy. Note that a distributional prior complements the ponder loss in guiding overall halting towards a target depth, but it cannot capture remainder information -- using ACT-agnostic distributional prior alone results in an accuracy drop of more than $2\%$.

\begin{table}[t]
\centering
\resizebox{.8\linewidth}{!}{
\begin{tabular}{lrrrr}
\toprule
\multirow{2}{*}{\textbf{Method}} & \multicolumn{2}{c}{\textbf{Efficiency}} & \multirow{2}{*}{\textbf{Top-1 Acc.}$\uparrow$} & \multirow{2}{*}{\textbf{Throughput}}  \\
\cmidrule{2-3} 
& Params. $\downarrow$ & FLOPs $\downarrow$ &  & \\
% \midrule
% Work a & - & TBD &  $TBD$ \\
% Work b & - & TBD &  $TBD$ \\
% Work c & - & TBD &  $TBD$ \\
% \textbf{Ours - B} & - & TBD &  training \\
\midrule
ViT-B~\cite{dosovitskiy2020image} & $86$M    & $17.6$G & $77.9$ & $0.3$K imgs/s \\
DeiT-S~\cite{touvron2021training} & $22$M     & $4.6$G & $78.9$ & $0.8$K  imgs/s\\
DynamicViT~\cite{rao2021dynamicvit}  & $23$M    & $3.4$G & $78.3$ & $1.0$K  imgs/s\\
% DynamicViT~\cite{rao2021dynamicvit} &   $22+$M    & $2.9$G  &  $77.4$ \\
\rowcolor{lgreen}
\textbf{A-ViT-S}   & $22$M & $3.6$G   &  $78.6$ & $1.1$K  imgs/s\\
\rowcolor{lgreen}
\textbf{A-ViT-S + distl.}   & $22$M & $3.6$G   &  $80.7$ & $1.1$K  imgs/s\\
\midrule
DeiT-T~\cite{touvron2021training}  & $5$M    & $1.2$G  & $71.3$ & $2.1$K  imgs/s\\
DynamicViT~\cite{rao2021dynamicvit}  & $5.9$M    & $0.9$G & $70.9$ & $2.9$K  imgs/s\\
% DynamicViT~\cite{rao2021dynamicvit}  &  $5.9$M & $0.91$G &  TBD \\
% \textbf{DynaViTe - T}   & $\mathbf{5}$M & $\mathbf{0.8}$G & $\mathbf{2.8}$K & tbd & $\mathbf{72.2}$ \\
\rowcolor{lgreen}
\textbf{A-ViT-T}   & $5$M & $0.8$G &  $71.0$ & $3.4$K imgs/s\\
\rowcolor{lgreen}
\textbf{A-ViT-T + distl.}   & $5$M & $0.8$G   &  $72.4$ & $3.4$K  imgs/s\\
\bottomrule
\end{tabular}}
\caption{Throughput improvement enabled via adaptive tokens. Models with \textbf{+ distil.} is augmented with distillation token.}
\vspace{-3mm}
\label{tab:latency}
\end{table}

\noindent\textbf{``Free'' embedding to learn halting.} Next we justify the usage of a single value in the embedding vector for halting score computation and representation. In the embedding vectors, we set one entry at a random index to zero and analyze the associated accuracy drop without any finetuning of the model. Repeating $10$ times 
% \PM{does it mean you selected only 5 elements in total? might be not enough}\yin{will make it 10 today}\PM{should we then say that we tested the first 5-10, it will be enough, as random might never catch the 0 element}
for DeiT-T/S variants, the ImageNet1K top-1 accuracy only drops by $0.08\% \pm 0.04\%$/$0.04\% \pm 0.03\%$, respectively. This experiment demonstrates that one element in the vector can be used for another task with minimal impact on the original performance. In our experiments, we pick the first element in the vector and use it for the halting score computation.
% that embedding possesses sufficient capacity to spare one position for free to learn halting. As an oracle to estimate performance degradation when ablating an embedding position away from the underlying task, 

\noindent\textbf{Layer-wise networks to learn halting.}
We continue to examine viability to leverage extra networks for halting learning. To this end we add an extra two-layer learnable network (with input/hidden dimensions of $192/96$, internal/output gates as \texttt{GeLU}/\texttt{Sigmoid}) on top of embeddings of each layer in A-ViT-T. We observed a very slight increase in accuracy of $+0.06\%$ with $+0.2$M parameter and $-12.6\%$ inference throughput overhead, as auxiliary nets have to be executed sequentially with ViT layers. Given this tradeoff, we base learning of halting on existing ViT parameters.

% This offers the design freedom to ablate one embedding position and its associated learn-able parameters as inputs to compute $h$.

% \subsection{Ablations.}
% \noindent\textbf{Token- vs. layer-wise ACT.} Here we want to give insights to the difference when moving from per-layer to per-token ACT. Finer granularity allows for enhanced dynamic discarding mechanism.

% \noindent\textbf{Varying encoding positions.} 
% We earlier argue that the position used for halting score calculation allows for freedom of choice as we used index $i=0$ for all sigmoid calculation. We now randomly sample this $i$ and observed very similar performance. 

% \noindent\textbf{Sparing tokens from transformers.} 
% We observe there is a drop of accuracy and that happens mainly because of mixup. 

% \noindent\textbf{Distributional prior regularization.} 
% With distributional prior is much easier to control the target threshold. We would like to have some ablations here too.

% \noindent\textbf{Accuracy drop source due to Mixup.} 
% We observe there is a drop of accuracy and that happens mainly because of mixup. 

\section{Limitations \& Future Directions}
In this work we primarily focused on the classification task. However, extension to other tasks such as video processing can be of great interest, given not only spatial but also temporal redundancy within input tokens. 

\section{Conclusions}
We have introduced A-ViT to adaptively adjust the amount of token computation based on input complexity. We demonstrated that the method improves vision transformer throughput on hardware without imposing extra parameters or modifications of transformer blocks, outperforming prior dynamic approaches. Captured token importance distribution adaptively varies by input images, yet coincides surprisingly well with human perception, offering insights for future work to improve vision transformer efficiency.

%%%%%%%%% REFERENCES
{
    % \clearpage
    \small
    \bibliographystyle{ieee_fullname}
    \bibliography{reference}
}

% --- supplementary material
\appendix

\clearpage
% --- PDF will be split by an editor (e.g. macOS preview), so need to restart from page 1
\setcounter{page}{1}

% % --- repeat the title (AT: haven't found a more elegant way to do this...)
% \twocolumn[
% \centering
% \Large
% \textbf{A-ViT: Adaptive Tokens for Efficient Vision Transformer} \\
% \vspace{0.5em} Supplementary Material \\
% \vspace{1.0em}
% ] %< twocolumn

% \appendix

% \section{Extra experiments}

% \section{Dataset description}

\begin{figure*}[b]

\begingroup
\renewcommand*{\arraystretch}{0.3}
\begin{tabular}{l}
\large{\textbf{Appendix A - More Examples}}
\vspace{3mm}
\end{tabular}
\endgroup

\centering

\resizebox{.84\linewidth}{!}{
\begingroup
\renewcommand*{\arraystretch}{0.3}
\begin{tabular}{ccc}%original
\includegraphics[width=0.25\linewidth]{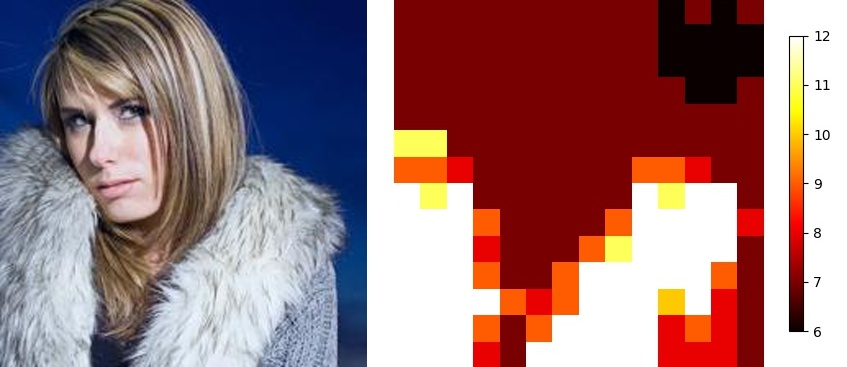} &
\includegraphics[width=0.25\linewidth]{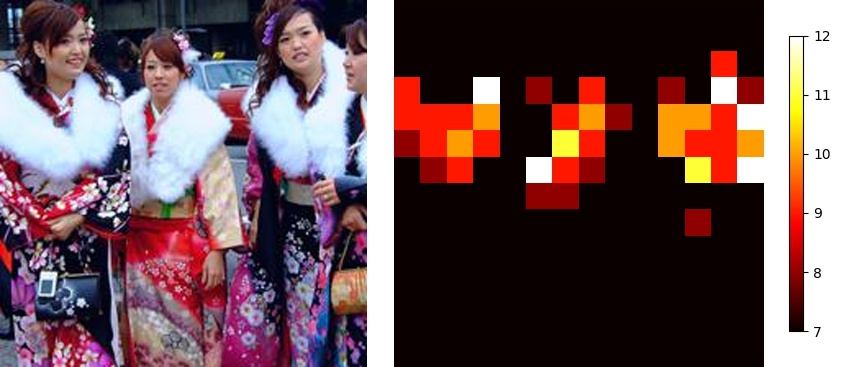} &
\includegraphics[width=0.25\linewidth]{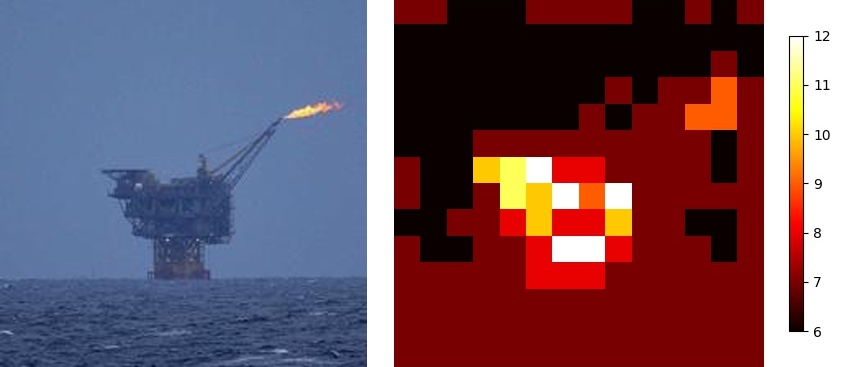} 
% & \includegraphics[width=0.25\linewidth]{fig/suppl/images_heat/z_class657_combined.jpg} 
\\
\scriptsize{fur coat} & \scriptsize{kimono} & \scriptsize{drilling platform}
\end{tabular}
\endgroup
}

% \vspace{1mm}

\resizebox{.84\linewidth}{!}{
\begingroup
\renewcommand*{\arraystretch}{0.3}
\begin{tabular}{ccc}%original
\includegraphics[width=0.25\linewidth]{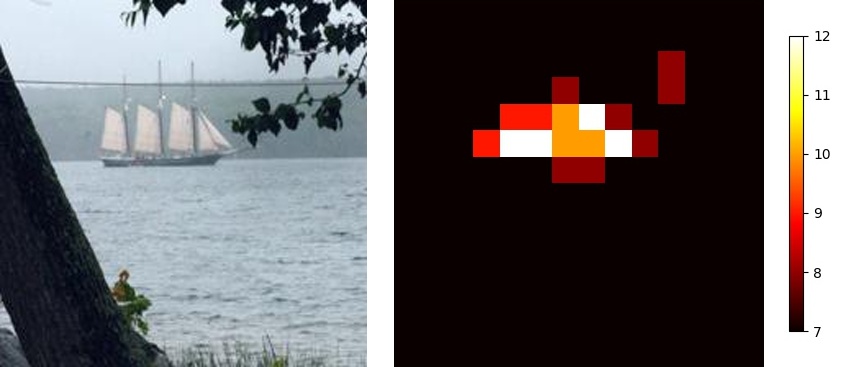} &
\includegraphics[width=0.25\linewidth]{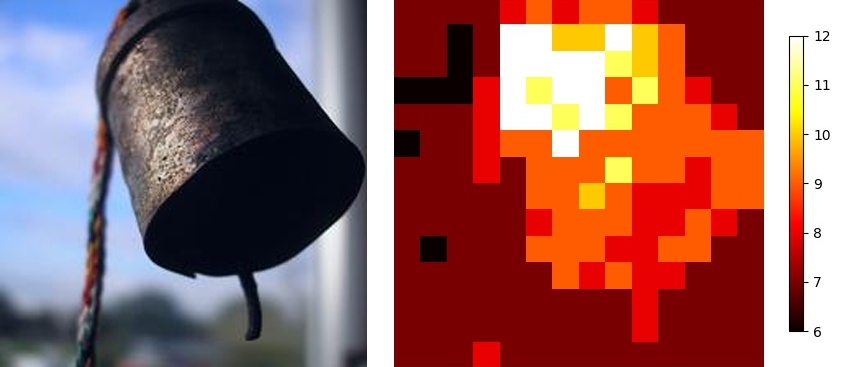} &
\includegraphics[width=0.25\linewidth]{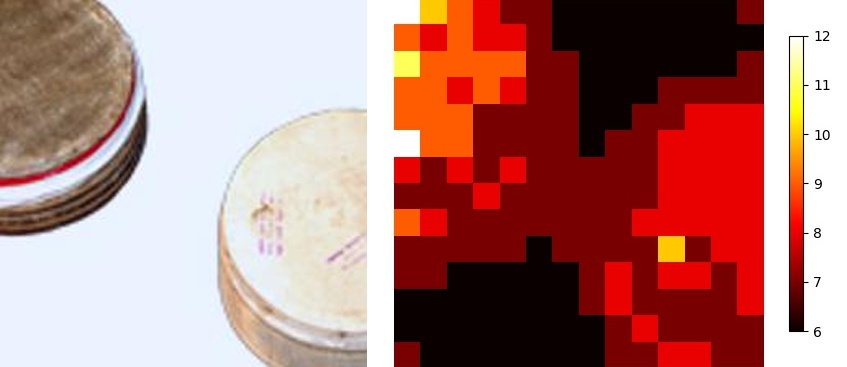} 
% & \includegraphics[width=0.25\linewidth]{fig/suppl/images_heat/z_class429_combined.jpg} 
\\

\scriptsize{schooner} & \scriptsize{bell} & \scriptsize{gong}

\end{tabular}
\endgroup
}

% \vspace{1mm}

\resizebox{.84\linewidth}{!}{
\begingroup
\renewcommand*{\arraystretch}{0.3}
\begin{tabular}{ccc}%original
\includegraphics[width=0.25\linewidth]{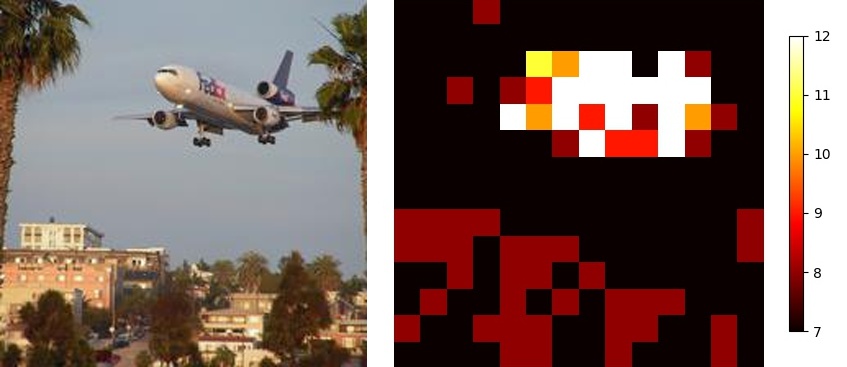} &
\includegraphics[width=0.25\linewidth]{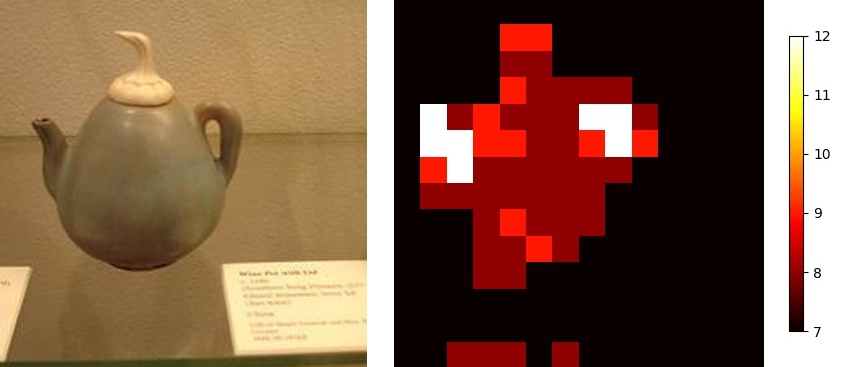} &
\includegraphics[width=0.25\linewidth]{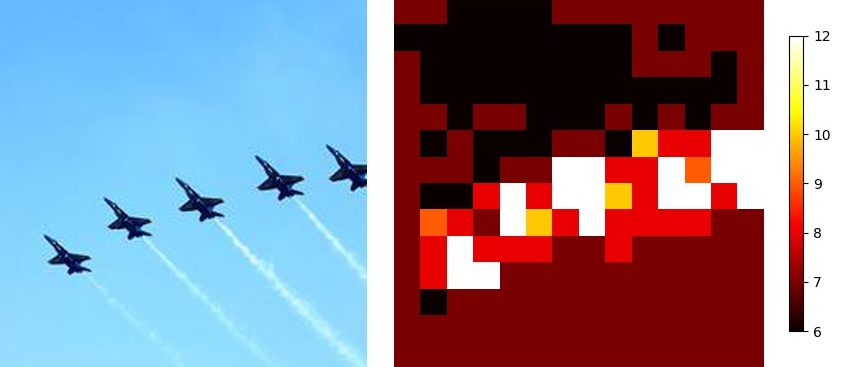} \\
% & \includegraphics[width=0.25\linewidth]{fig/suppl/images_heat/z_class389_combined.jpg}

\scriptsize{airliner} & \scriptsize{pitcher} & \scriptsize{attack aircraft carrier}

\end{tabular}
\endgroup
}

\vspace{1mm}

\resizebox{.84\linewidth}{!}{
\begingroup
\renewcommand*{\arraystretch}{0.3}
\begin{tabular}{ccc}%original
\includegraphics[width=0.25\linewidth]{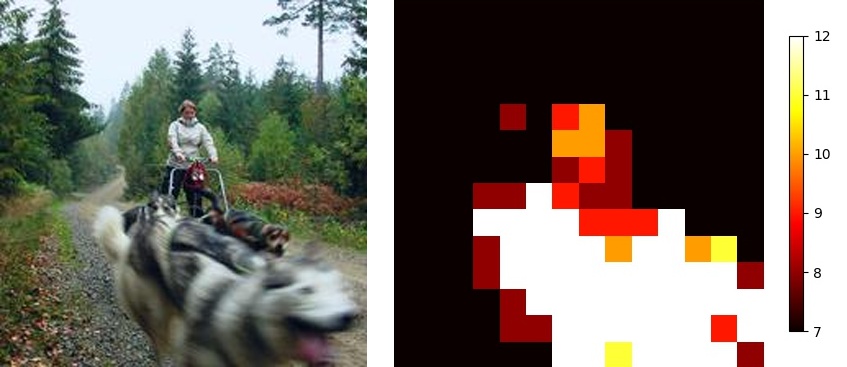} &
\includegraphics[width=0.25\linewidth]{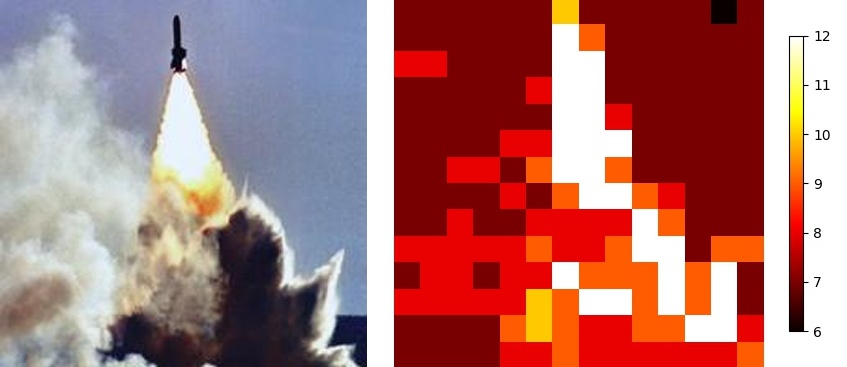} &
\includegraphics[width=0.25\linewidth]{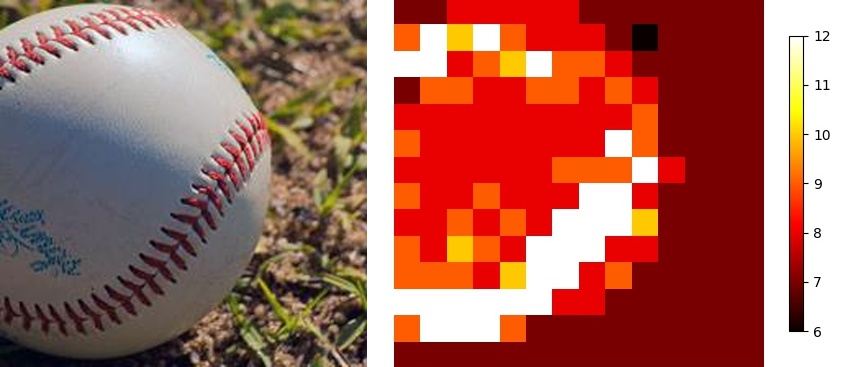} 
% & \includegraphics[width=0.25\linewidth]{fig/suppl/images_heat/z_class526_combined.jpg} 
\\

\scriptsize{dogsled} & \scriptsize{missile} & \scriptsize{baseball}

\end{tabular}
\endgroup
}

% \vspace{1mm}

\resizebox{.84\linewidth}{!}{
\begingroup
\renewcommand*{\arraystretch}{0.3}
\begin{tabular}{ccc}%original
\includegraphics[width=0.25\linewidth]{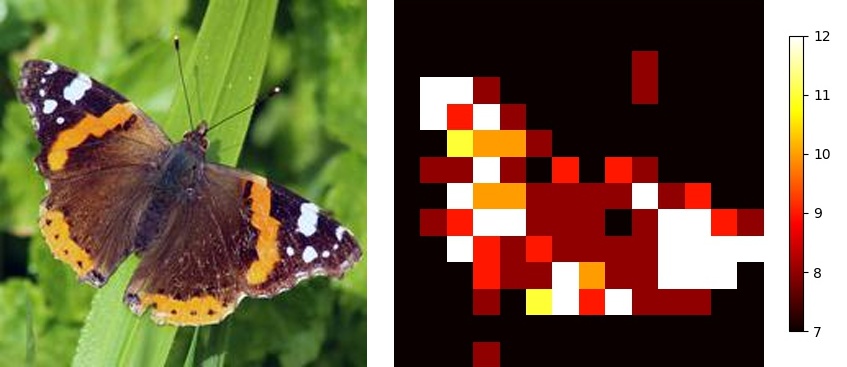} &
\includegraphics[width=0.25\linewidth]{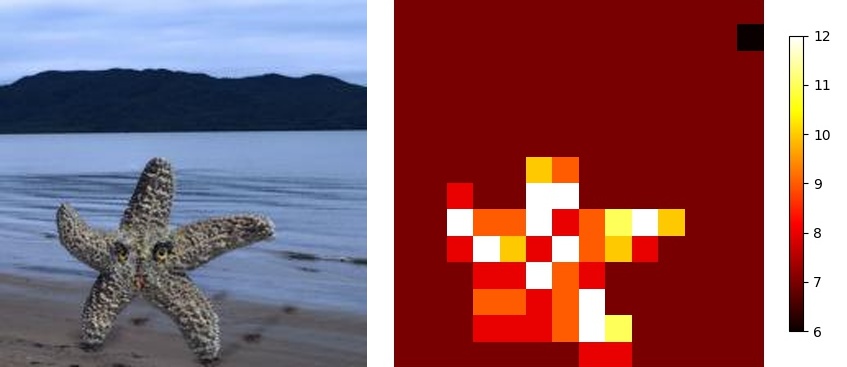} &
\includegraphics[width=0.25\linewidth]{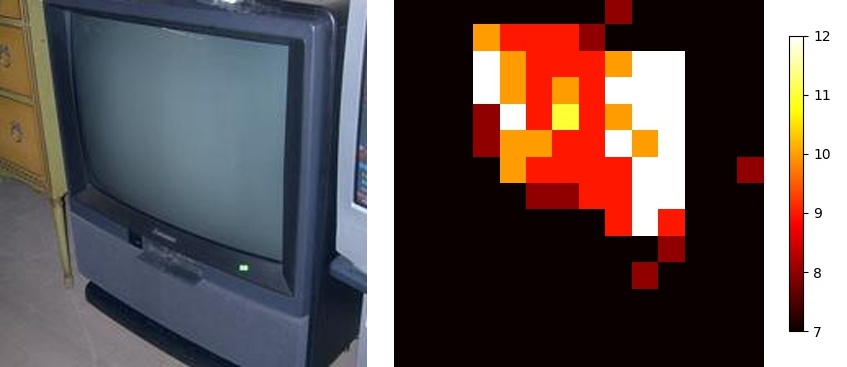} 
% & \includegraphics[width=0.25\linewidth]{fig/suppl/images_heat/z_class288_combined.jpg} 
\\

\scriptsize{admiral butterfly} & \scriptsize{starfish} & \scriptsize{screen}

\end{tabular}
\endgroup
}

% \vspace{1mm}

\resizebox{.84\linewidth}{!}{
\begingroup
\renewcommand*{\arraystretch}{0.3}
\begin{tabular}{ccc}%original
\includegraphics[width=0.25\linewidth]{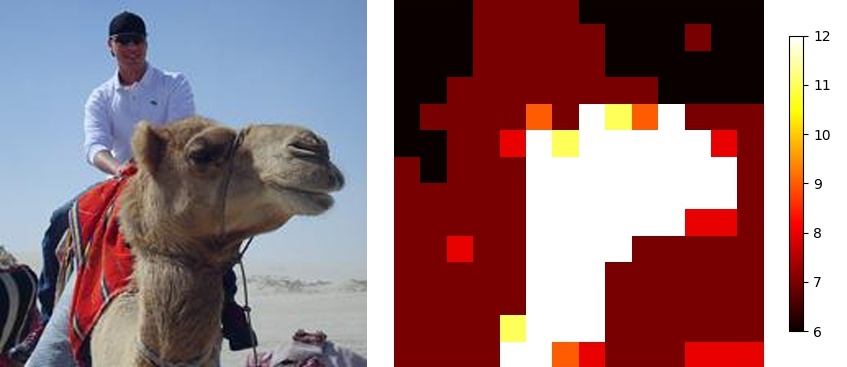} &
\includegraphics[width=0.25\linewidth]{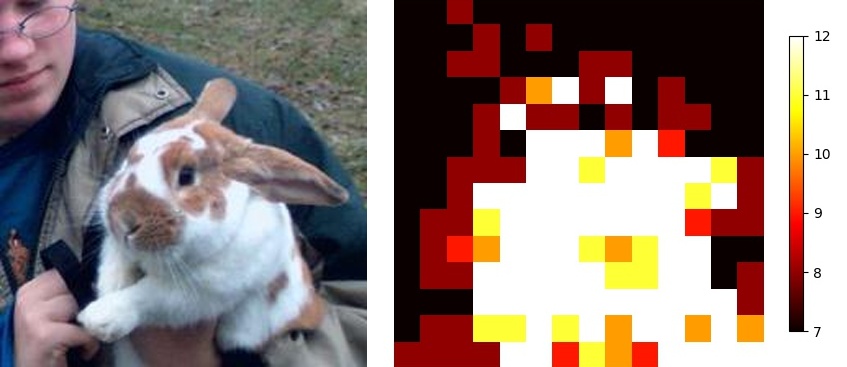} &
\includegraphics[width=0.25\linewidth]{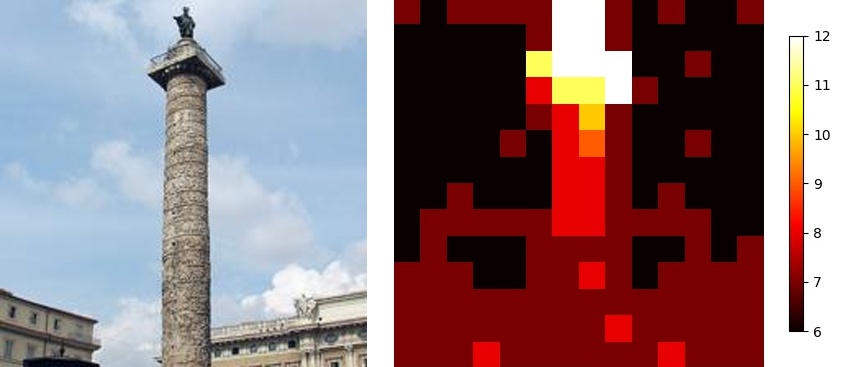} 
% & \includegraphics[width=0.25\linewidth]{fig/suppl/images_heat/z_class110_combined.jpg} 
\\

\scriptsize{Arabian camel} & \scriptsize{wood rabbit} & \scriptsize{obelisk}

\end{tabular}
\endgroup
}

% \vspace{1mm}

\resizebox{.84\linewidth}{!}{
\begingroup
\renewcommand*{\arraystretch}{0.3}
\begin{tabular}{ccc}%original
\includegraphics[width=0.25\linewidth]{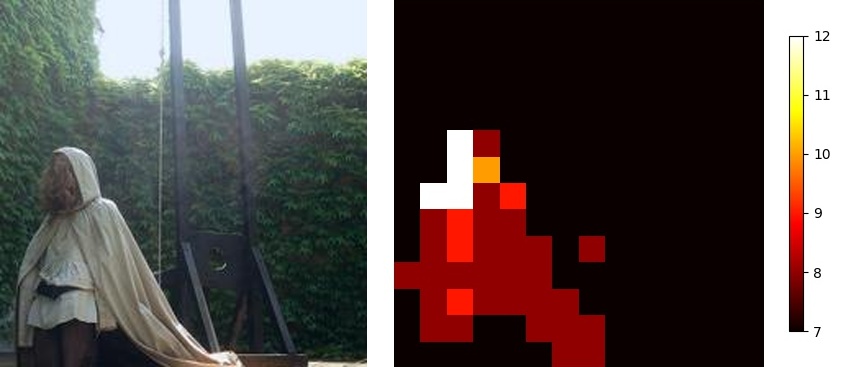} &
\includegraphics[width=0.25\linewidth]{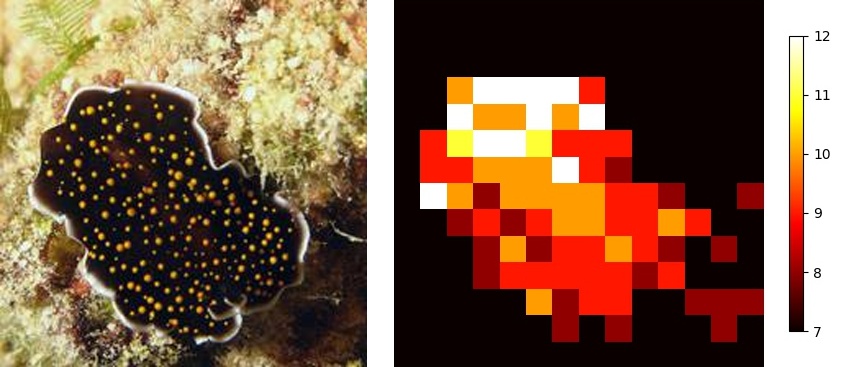} &
\includegraphics[width=0.25\linewidth]{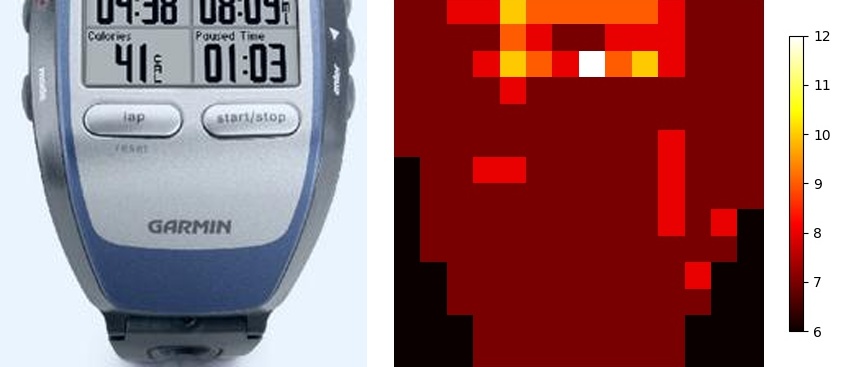} 
% & \includegraphics[width=0.25\linewidth]{fig/suppl/images_heat/z_class110_combined.jpg} 
\\
\scriptsize{guillotine} & \scriptsize{platyhelminth} & \scriptsize{digital watch}
\end{tabular}
\endgroup
}

% \vspace{1mm}

\resizebox{.84\linewidth}{!}{
\begingroup
\renewcommand*{\arraystretch}{0.3}
\begin{tabular}{ccc}%original
\includegraphics[width=0.25\linewidth]{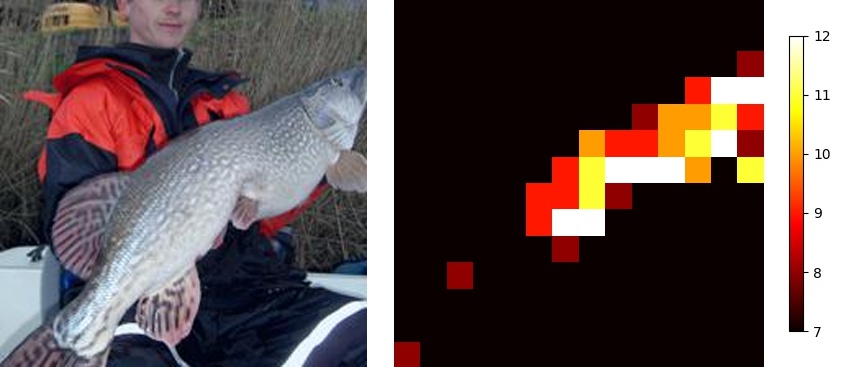} &
\includegraphics[width=0.25\linewidth]{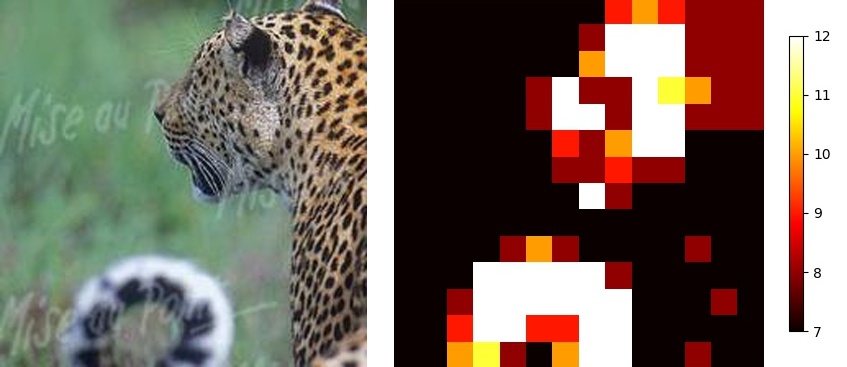} &
\includegraphics[width=0.25\linewidth]{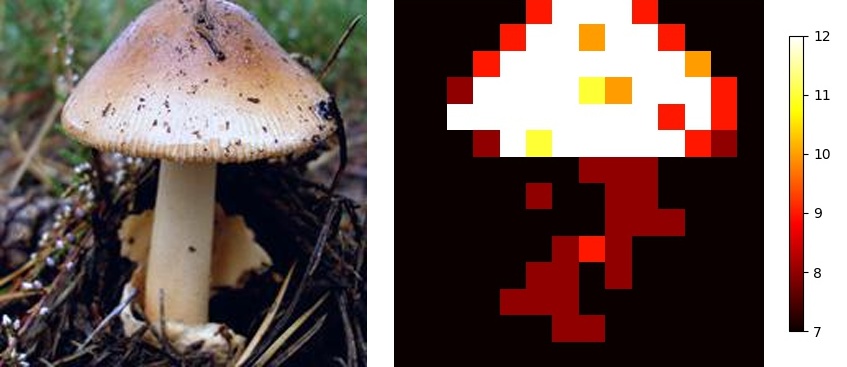} 
% & \includegraphics[width=0.25\linewidth]{fig/suppl/images_heat/z_class110_combined.jpg} 
\\
\scriptsize{barracouta} & \scriptsize{leopard} & \scriptsize{mushroom}
\end{tabular}
\endgroup
}

% \vspace{1mm}

\resizebox{.84\linewidth}{!}{
\begingroup
\renewcommand*{\arraystretch}{0.3}
\begin{tabular}{ccc}%original
\includegraphics[width=0.25\linewidth]{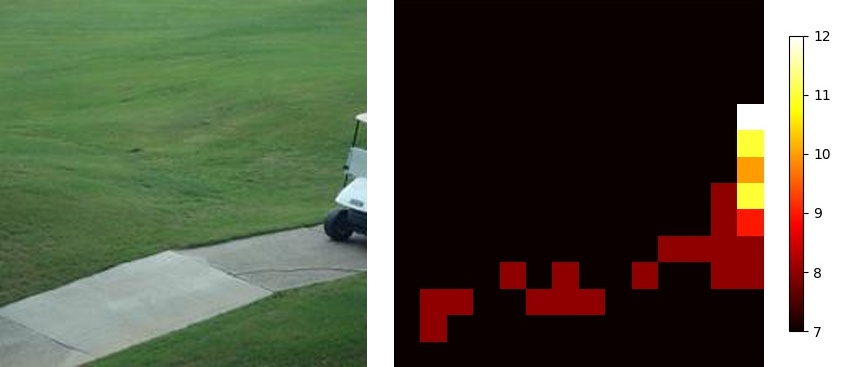} &
\includegraphics[width=0.25\linewidth]{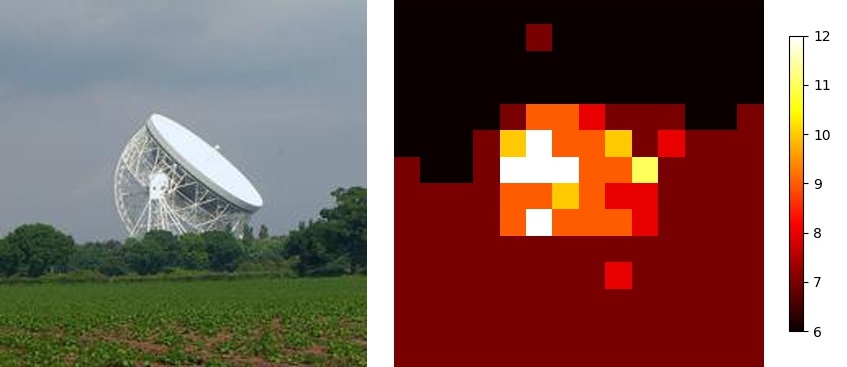} &
\includegraphics[width=0.25\linewidth]{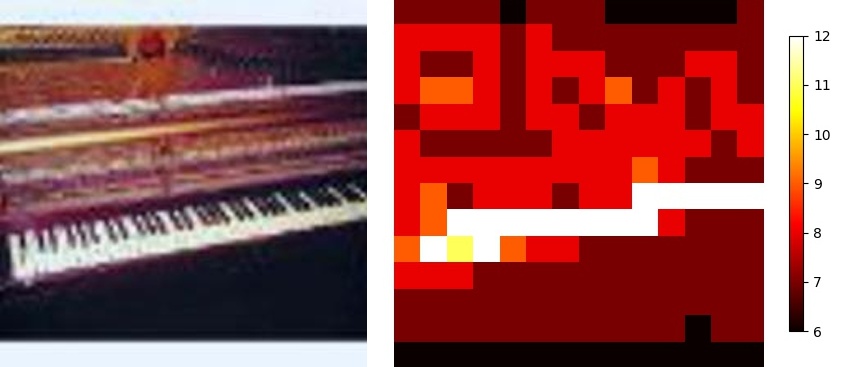} 
% & \includegraphics[width=0.25\linewidth]{fig/suppl/images_heat/z_class110_combined.jpg} 
\\
\scriptsize{golf-cart} & \scriptsize{radio telescope} & \scriptsize{upright piano}
\end{tabular}
\endgroup
}

% \vspace{1mm}

% \caption{More examples of across a more diverse set of image categories -- original image (left) and the dynamic token depth (right) of A-ViT-T on the ImageNet-1K validation set. Best viewed in color.}

\caption{
Additional examples across a more diverse set of image categories -- original image (left) and the dynamic token depth (right) of A-ViT-T on the ImageNet-1K validation set. Again adaptive tokens can quickly cater to informative regions while filtering out complex backgrounds, \eg, completely ignoring human faces and focusing on the coats, see the first two image on the first row. Even for a very small informative region of the target object, the computation can still be effectively allocated towards it, see the first golf-cart class sample of the last row as an example.
}

\label{fig:supp_all_images}
\end{figure*}

\clearpage

\section*{Appendix B - Additional Details}

% \section{More Examples}
% We include more visual examples covering a more diverse set of image categories in Fig.~\ref{fig:supp_all_images}. Again adaptive tokens can quickly cater to informative regions while filtering out complex backgrounds, \eg, completely ignoring human faces and focusing on the collars and coats, see the first two images on the first row. Even for a very small informative region of the target object, the computation can still be effectively allocated towards the region, see the first golf-cart class sample of the last row as an example.

% \section{Experimental Details}

\paragraph{Training} For training setup other than the scaling constants, \texttt{lr} specified in the main manuscript, we follow original repository for all other hyper-parameters at \url{https://github.com/facebookresearch/DeiT} such as drop out rate, momentum, preprocessing, etc, imposing minimum training recipe changes when adapting a static model to its adaptive counterpart.

\paragraph{Latency} We measure the latency on an NVIDIA TITAN RTX $2080$ GPU with PyTorch for batch size of $64$ images, CUDA $10.2$. For GPU warming up, $100$ forward passes are conducted, and then the median speed of the $1$K measurements of the full model latency are reported. The exact same setup is shared across all baseline and proposed methods for a fair comparison.

\paragraph{SOTA baselines.} We followed DeiT's~[43] repository\footnote{\texttt{\scriptsize{https://github.com/facebookresearch/DeiT}}} for recipes and checkpoints as a common starting point for all experiments. For DynamicViT~\cite{rao2021dynamicvit}, we used the public repository and script from the authors. For other dynamic approaches from CNN/NLP literature, we re-implemented the methods on DeiT to examine ACT~\cite{graves2016adaptive} for layer-wise halting, confidence threshold~\cite{weijie2020fastbert} on post-softmax logits, two variants of similarity gauging~\cite{elbayad2019depth} on delta-logits based on (i) LPIPS and (ii) MSE similarity scores, and PonderNet~\cite{banino2021pondernet} with geometric-distribution sampling towards token halting. For all methods, a detailed grid search was conducted to ensure optimal hyper-parameters. 

% --- uncomment this to read the instructions
% \input{sec/X_instructions}

\end{document}